\documentclass[10pt,twocolumn,letterpaper]{article}

\usepackage{cvpr}
\usepackage{times}
\usepackage{epsfig}
\usepackage{graphicx}
\usepackage{amsmath}
\usepackage{amssymb}
\usepackage{rotating}
\usepackage{color}
\usepackage{framed}
\usepackage{subcaption}
\usepackage{balance}
\usepackage{booktabs}
\usepackage{bm}
\usepackage{bbm}
\usepackage{enumitem}
\usepackage[export]{adjustbox}
\usepackage{pifont}
\usepackage{fancyvrb}
\usepackage{lipsum}

\usepackage[pagebackref=true,breaklinks=true,colorlinks,bookmarks=false]{hyperref}
\definecolor{darkgreen}{rgb}{0.03, 0.58, 0.31}
\newcommand{\tit}[1]{\smallbreak\noindent\textbf{#1.}}
\newcommand{\tinytit}[1]{\noindent\textbf{#1.}}

\def \ours {$\mathcal{M}^2$ Transformer\xspace}

\newcommand\blfootnote[1]{%
  \begingroup
  \renewcommand\thefootnote{}\footnote{#1}%
  \addtocounter{footnote}{-1}%
  \endgroup
}

\cvprfinalcopy 

\def\cvprPaperID{***} 

\ifcvprfinal\fi
\begin{document}

\title{Meshed-Memory Transformer for Image Captioning}

\author{Marcella Cornia$^*$ \quad Matteo Stefanini$^*$ \quad Lorenzo Baraldi$^*$ \quad Rita Cucchiara \\
University of Modena and Reggio Emilia \\
{\tt\small \{name.surname\}@unimore.it}
}

\maketitle
\thispagestyle{empty}

\begin{abstract}
Transformer-based architectures represent the state of the art in sequence modeling tasks like machine translation and language understanding. Their applicability to multi-modal contexts like image captioning, however, is still largely under-explored.
With the aim of filling this gap, we present $\mathcal{M}^2$ -- a Meshed Transformer with Memory for Image Captioning. The architecture improves both the image encoding and the language generation steps: it learns a multi-level representation of the relationships between image regions integrating learned a priori knowledge, and uses a mesh-like connectivity at decoding stage to exploit low- and high-level features. 
Experimentally, we investigate the performance of the $\mathcal{M}^2$ Transformer and different fully-attentive models in comparison with recurrent ones. When tested on COCO, our proposal achieves a new state of the art in single-model and ensemble configurations on the ``Karpathy'' test split and on the online test server. We also assess its performances when describing objects unseen in the training set.
Trained models and code for reproducing the experiments are publicly available at: \url{https://github.com/aimagelab/meshed-memory-transformer}.  
\blfootnote{$^*$Equal contribution.}
\end{abstract}

\section{Introduction}
Image captioning is the task of describing the visual content of an image in natural language. As such, it requires an algorithm to understand and model the relationships between visual and textual elements, and to generate a sequence of output words. This has usually been tackled via Recurrent Neural Network models~\cite{vinyals2015show,karpathy2015deep,xu2015show,vinyals2017show,cornia2019show}, in which the sequential nature of language is modeled with the recurrent relations of either RNNs or LSTMs. Additive attention or graph-like structures~\cite{yao2018exploring} are often added to the recurrence~\cite{xu2015show,huang2019attention} in order to model the relationships between image regions, words, and eventually tags~\cite{liu2019entangled}.

\begin{figure}[t]
    \centering
    \includegraphics[width=0.99\columnwidth]{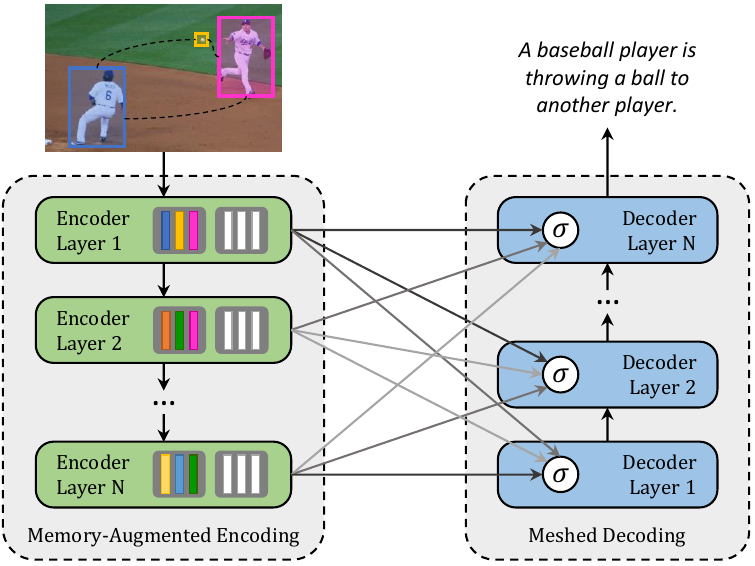}
    \caption{Our image captioning approach encodes relationships between image regions exploiting learned a priori knowledge. Multi-level encodings of image regions are connected to a language decoder through a meshed and learnable connectivity.}
    \label{fig:first_page}
    \vspace{-.3cm}
\end{figure}

This schema has remained the dominant approach in the last few years, with the exception of the investigation of Convolutional language models~\cite{aneja2018convolutional}, which however did not become a leading choice. The recent advent of fully-attentive models, in which the recurrent relation is abandoned in favour of the use of self-attention, offers unique opportunities in terms of set and sequence modeling performances, as testified by the Transformer~\cite{vaswani2017attention} and BERT~\cite{devlin2018bert} models and their applications to retrieval~\cite{song2019polysemous} and video understanding~\cite{sun2019videobert}. Also, this setting offers novel architectural modeling capabilities, as for the first time the attention operator is used in a multi-layer and extensible fashion. Nevertheless, the multi-modal nature of image captioning demands for specific architectures, different from those employed for the understanding of a single modality.

Following this premise, we investigate the design of a novel fully-attentive approach for image captioning. Our architecture takes inspiration from the Transformer model~\cite{vaswani2017attention} for machine translation and incorporates two key novelties with respect to all previous image captioning algorithms: (\textit{i})~image regions and their relationships are encoded in a multi-level fashion, in which low-level and high-level relations are taken into account. When modeling these relationships, our model can learn and encode a priori knowledge by using persistent \textit{memory vectors}. (\textit{ii})~The generation of the sentence, done with a multi-layer architecture, exploits both low- and high-level visual relationships instead of having just a single input from the visual modality. This is achieved through a learned gating mechanism, which weights multi-level contributions at each stage. As this creates a mesh connectivity schema between encoder and decoder layers, we name our model \textit{Meshed-Memory Transformer} -- $\mathcal{M}^2$ Transformer for short. Figure~\ref{fig:first_page} depicts a schema of the architecture.

Experimentally, we explore different fully-attentive baselines and recent proposals, gaining insights on the performance of fully-attentive models in image captioning. Our $\mathcal{M}^2$ Transformer, when tested on the COCO benchmark, achieves a new state of the art on the ``Karpathy'' test set, on both single-model and ensemble configurations. Most importantly, it surpasses existing proposals on the online test server, \textit{ranking first among published algorithms}.

\tit{Contributions} To sum up, our contributions are as follows:
\begin{itemize}[noitemsep,topsep=0pt]
    \item We propose a novel fully-attentive image captioning algorithm. Our model encapsulates a multi-layer encoder for image regions and a multi-layer decoder which generates the output sentence. To exploit both low-level and high-level contributions, encoding and decoding layers are connected in a mesh-like structure, weighted through a learnable gating mechanism;
    \item In our visual encoder, relationships between image regions are encoded in a multi-level fashion exploiting learned a priori knowledge, which is modeled via persistent memory vectors;
    \item  We show that the $\mathcal{M}^2$ Transformer surpasses all previous proposals for image captioning, achieving a new state of the art on the online COCO evaluation server;
    \item As a complementary contribution, we conduct experiments to compare different fully-attentive architectures on image captioning and validate the performance of our model on novel object captioning, using the recently proposed nocaps dataset. Finally, to improve reproducibility and foster new research in the field, we will publicly release the source code and trained models of all experiments.
\end{itemize}

\noindent

\section{Related work}
A broad collection of methods have been proposed in the field of image captioning in the last few years.
Earlier captioning approaches were based on the generation of simple templates, filled by the output of an object detector or attribute predictor~\cite{socher2010connecting,yao2010i2t}.
With the advent of Deep Neural Networks, most captioning techniques have employed RNNs as language models and used the output of one or more layers of a CNN to encode visual information and condition language generation~\cite{vinyals2016show,rennie2017self,donahue2015long,johnson2016densecap}.
On the training side, while initial methods were based on a time-wise cross-entropy training, a notable achievement has been made with the introduction of Reinforcement Learning, which enabled the use of non-differentiable caption metrics as optimization objectives~\cite{rennie2017self,ranzato2015sequence,liu2017improved}.
On the image encoding side, instead, single-layer attention mechanisms have been adopted to incorporate spatial knowledge, initially from a grid of CNN features~\cite{xu2015show,lu2017knowing,you2016image}, and then using image regions extracted with an object detector~\cite{anderson2018bottom,pedersoli2017areas,lu2018neural}.
To further improve the encoding of objects and their relationships, Yao~\etal~\cite{yao2018exploring} have proposed to use a graph convolution neural network in the image encoding phase to integrate semantic and spatial relationships between objects.
On the same line, Yang \etal~\cite{yang2019auto} used a multi-modal graph convolution network to modulate scene graphs into visual representations.

Despite their wide adoption, RNN-based models suffer from their limited representation power and sequential nature. After the emergence of Convolutional language models, which have been explored for captioning as well~\cite{aneja2018convolutional},
new fully-attentive paradigms~\cite{vaswani2017attention,devlin2018bert,sukhbaatar2019augmenting} have been proposed and achieved state-of-the-art results in machine translation and language understanding tasks. Likewise, some recent approaches have investigated the application of the Transformer model~\cite{vaswani2017attention} to the image captioning task.

In a nutshell, the Transformer comprises an encoder made of a stack of self-attention and feed-forward layers, and a decoder which uses self-attention on words and cross-attention over the output of the last encoder layer. Herdade~\etal~\cite{herdade2019image} used the Transformer architecture for image captioning and incorporated geometric relations between detected input objects. In particular, they computed an additional geometric weight between object pairs which is used to scale attention weights.
Li~\etal~\cite{liu2019entangled} used the Transformer in a model that exploits visual information and additional semantic knowledge given by an external tagger. On a related line, Huang~\etal~\cite{huang2019attention} introduced an extension of the attention operator in which the final attended information is weighted by a gate guided by the context. In their approach, a Transformer-like encoder was paired with an LSTM decoder.
While the aforementioned approaches have exploited the original Transformer architecture, in this paper we devise a novel fully-attentive model that improves the design of both the image encoder and the language decoder, introducing two novel attention operators and a different design of the connectivity between encoder and decoder.

\begin{figure*}[t]
    \centering
    \includegraphics[width=\linewidth]{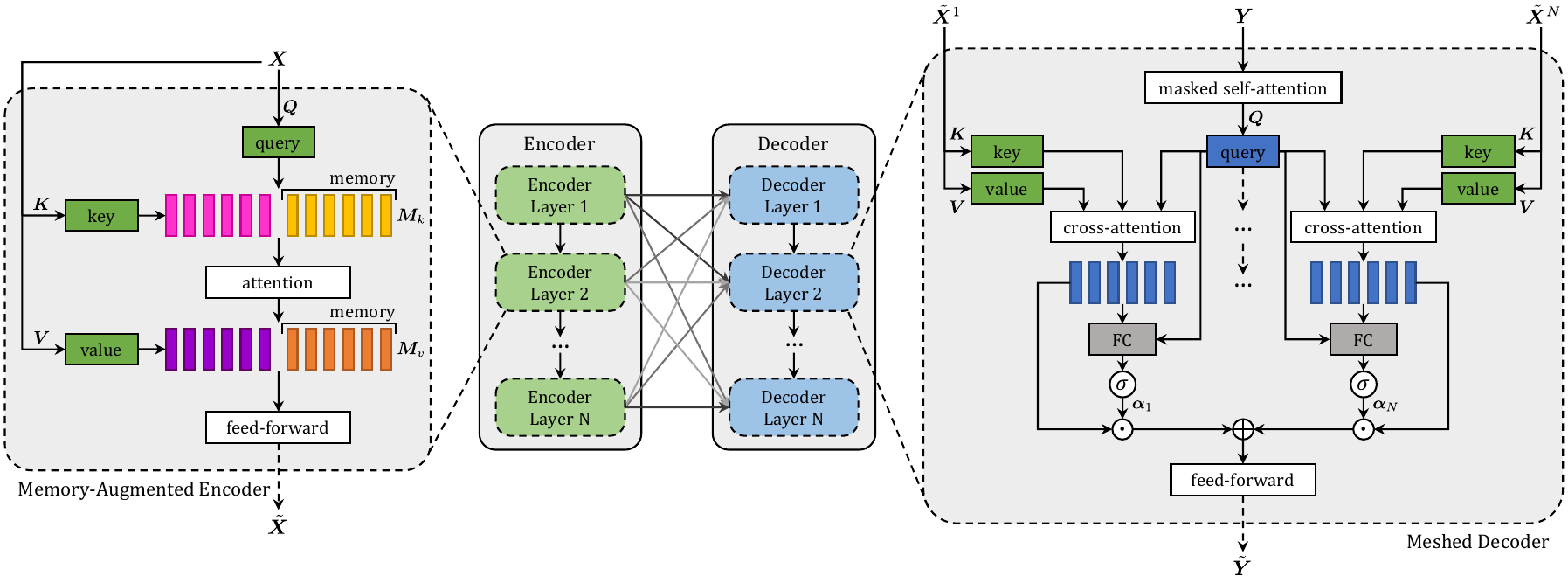}
    \caption{Architecture of the $\mathcal{M}^2$ Transformer. Our model is composed of a stack of memory-augmented encoding layers, which encodes multi-level visual relationships with a priori knowledge, and a stack of decoder layers, in charge of generating textual tokens. For the sake of clarity, $\mathsf{AddNorm}$ operations are not shown. Best seen in color.}
    \label{fig:method}
    \vspace{-.3cm}
\end{figure*}

\newcommand{\selfatt}{\mathcal{S}}
\newcommand{\crossatt}{\mathcal{C}}
\newcommand{\memoryatt}{\mathcal{M}_\text{mem}}
\newcommand{\meshatt}{\mathcal{M}_\text{mesh}}

\section{Meshed-Memory Transformer}
Our model can be conceptually divided into an encoder and a decoder module, both made of stacks of attentive layers. While the encoder is in charge of processing regions from the input image and devising relationships between them, the decoder reads from the output of each encoding layer to generate the output caption word by word.
All intra-modality and cross-modality interactions between word and image-level features are modeled via scaled dot-product attention, without using recurrence. Attention operates on three sets of vectors, namely a set of queries $\bm{Q}$, keys $\bm{K}$ and values $\bm{V}$, and takes a weighted sum of value vectors according to a similarity distribution between query and key vectors. In the case of scaled dot-product attention, the operator can be formally defined as
\begin{align}
\mathsf{Attention}(\bm{Q}, \bm{K}, \bm{V})=\operatorname{softmax}\left(\frac{\bm{Q} \bm{K}^{T}}{\sqrt{d}}\right) \bm{V},
\label{eq:attention}
\end{align}
where $\bm{Q}$ is a matrix of $n_q$ query vectors, $\bm{K}$ and $\bm{V}$ both contain $n_k$ keys and values, all with the same dimensionality, and $d$ is a scaling factor.

\subsection{Memory-Augmented Encoder}
Given a set of image regions $\bm{X}$ extracted from an input image, attention can be used to obtain a permutation invariant encoding of $\bm{X}$ through the self-attention operations used in the Transformer~\cite{vaswani2017attention}. In this case, queries, keys, and values are obtained by linearly projecting the input features, and the operator can be defined as 
\begin{equation}
    \selfatt(\bm{X}) = \mathsf{Attention}(W_q\bm{X}, W_k\bm{X}, W_v\bm{X}),
\end{equation}
where $W_q, W_k, W_v$ are matrices of learnable weights. 
The output of the self-attention operator is a new set of elements $\selfatt(\bm{X})$, with the same cardinality as $\bm{X}$, in which each element of $\bm{X}$ is replaced with a weighted sum of the values, \ie~of linear projections of the input (Eq.~\ref{eq:attention}). 

Noticeably, attentive weights depend solely on the pairwise similarities between linear projections of the input set itself. Therefore, the self-attention operator can be seen as a way of encoding pairwise relationships inside the input set. When using image regions (or features derived from image regions) as the input set, $\selfatt(\cdot)$ can naturally encode the pairwise relationships between regions that are needed to understand the input image before describing it\footnote{Taking another perspective, self-attention is also conceptually equivalent to an attentive encoding of graph nodes~\cite{velivckovic2017graph}.}.

This peculiarity in the definition of self-attention has, however, a significant limitation. Because everything depends solely on pairwise similarities, self-attention cannot model a priori knowledge on relationships between image regions. For example, given one region encoding a man and a region encoding a basketball ball, it would be difficult to infer the concept of \textit{player} or \textit{game} without any a priori knowledge. Again, given regions encoding eggs and toasts, the knowledge that the picture depicts a \textit{breakfast} could be easily inferred using a priori knowledge on relationships.

\tinytit{Memory-Augmented Attention}
To overcome this limitation of self-attention, we propose a memory-augmented attention operator. In our proposal, the set of keys and values used for self-attention is extended with additional ``slots'' which can encode a priori information. To stress that a priori information should not depend on the input set $\bm{X}$, the additional keys and values are implemented as plain learnable vectors which can be directly updated via SGD. Formally, the operator is defined as:
\begin{gather}
    \memoryatt(\bm{X}) = \mathsf{Attention}(W_q\bm{X}, \bm{K}, \bm{V})\nonumber \\    
    \bm{K} = \left[ W_k \bm{X}, \bm{M}_k \right] \nonumber \\
    \bm{V} = \left[ W_v \bm{X}, \bm{M}_v \right],     
\end{gather}
where $\bm{M}_k$ and $\bm{M}_v$ are learnable matrices with $n_m$ rows, and $\left[ \cdot, \cdot \right]$ indicates concatenation. Intuitively, by adding learnable keys and values, through attention it will be possible to retrieve learned knowledge which is not already embedded in $\bm{X}$. At the same time, our formulation leaves the set of queries unaltered. 

Just like the self-attention operator, our memory-augmented attention can be applied in a multi-head fashion. In this case, the memory-augmented attention operation is repeated $h$ times, using different projection matrices $W_q, W_k, W_v$ and different learnable memory slots $\bm{M}_k, \bm{M}_v$ for each head. Then, we concatenate the results from different heads and apply a linear projection. 

\tinytit{Encoding layer} 
We embed our memory-augmented operator into a Transformer-like layer: the output of the memory-augmented attention is applied to a position-wise feed-forward layer composed of two affine transformations with a single non-linearity, which are independently applied to each element of the set. Formally,
\begin{align}
    \mathcal{F}(\bm{X})_i = U \sigma(V \bm{X}_i + b) + c,
    \label{eq:pwff}
\end{align}
where $\bm{X}_i$ indicates the $i$-th vector of the input set, and $\mathcal{F}(\bm{X})_i$ the $i$-th vector of the output. Also, $\sigma(\cdot)$ is the ReLU activation function, $V$ and $U$ are learnable weight matrices, $b$ and $c$ are bias terms.

Each of these sub-components (memory-augmented attention and position-wise feed-forward) is then encapsulated within a residual connection and a layer norm operation. The complete definition of an encoding layer can be finally written as:
\begin{align}
    \bm{Z} &= \mathsf{AddNorm}(\memoryatt(\bm{X})) \nonumber \\
    \bm{\tilde{X}} &= \mathsf{AddNorm}(\mathcal{F}(\bm{Z})),
\end{align}
where $\mathsf{AddNorm}$ indicates the composition of a residual connection and of a layer normalization.

\tinytit{Full encoder}
Given the aforementioned structure, multiple encoding layers are stacked in sequence, so that the $i$-th layer consumes the output set computed by layer $i-1$. This amounts to creating multi-level encodings of the relationships between image regions, in which higher encoding layers can exploit and refine relationships already identified by previous layers, eventually using a priori knowledge. A stack of $N$ encoding layers will therefore produce a multi-level output $\mathcal{\tilde{X}} = ( \bm{\tilde{X}}^1, ..., \bm{\tilde{X}}^N )$, obtained from the outputs of each encoding layer.

\subsection{Meshed Decoder}
Our decoder is conditioned on both previously generated words and region encodings, and is in charge of generating the next tokens of the output caption. Here, we exploit the aforementioned multi-level representation of the input image while still building a multi-layer structure. To this aim, we devise a meshed attention operator which, unlike the cross-attention operator of the Transformer, can take advantage of all encoding layers during the generation of the sentence. 

\tinytit{Meshed Cross-Attention}
Given an input sequence of vectors $\bm{Y}$, and outputs from all encoding layers $\mathcal{\tilde{X}}$, the Meshed Attention operator connects $\bm{Y}$ to all elements in $\mathcal{\tilde{X}}$ through gated cross-attentions. Instead of attending only the last encoding layer, we perform a cross-attention with all encoding layers. These multi-level contributions are then summed together after being modulated.
Formally, our meshed attention operator is defined as
\begin{equation}
    \meshatt(\mathcal{\tilde{X}}, \bm{Y}) = \sum_{i=1}^N \bm{\alpha}_i \odot \crossatt(\bm{\tilde{X}}^i, \bm{Y}),
    \label{eq:mesh_attention}
\end{equation}
where $\crossatt(\cdot, \cdot)$ stands for the encoder-decoder cross-attention, computed using queries from the decoder and keys and values from the encoder:
\begin{equation}
        \crossatt(\bm{\tilde{X}}^i, \bm{Y}) =  \mathsf{Attention}(W_q\bm{Y}, W_k\bm{\tilde{X}}^i, W_v\bm{\tilde{X}}^i),
\end{equation}
and $\bm{\alpha}_i$ is a matrix of weights having the same size as the cross-attention results.
Weights in $\bm{\alpha}_i$ modulate both the single contribution of each encoding layer, and the relative importance between different layers. These are computed by measuring the relevance between the result of the cross-attention computed with each encoding layer and the input query, as follows: 
\begin{equation}
    \bm{\alpha}_i = \sigma\left(W_i \left[ \bm{Y}, \crossatt(\bm{\tilde{X}}^i, \bm{Y}) \right] + b_i \right),
\end{equation}
where $\left[\cdot, \cdot\right]$ indicates concatenation, $\sigma$ is the sigmoid activation, $W_i$ is a $2d\times d$ weight matrix, and $b_i$ is a learnable bias vector.

\tinytit{Architecture of decoding layers}
As for encoding layers, we apply our meshed attention in a multi-head fashion. As the prediction of a word should only depend on previously predicted words, the decoder layer comprises a masked self-attention operation which connects queries derived from the $t$-th element of its input sequence $\bm{Y}$ with keys and values obtained from the left-hand subsequence, \ie~$\bm{Y}_{\le t}$. Also, the decoder layer contains a position-wise feed-forward layer (as in Eq.~\ref{eq:pwff}), and all components are encapsulated within $\mathsf{AddNorm}$ operations. The final structure of the decoder layer can be written as:
\begin{align}
    \bm{Z} &= \mathsf{AddNorm}(\meshatt(\mathcal{\tilde{X}}, \mathsf{AddNorm}(\selfatt_{\text{mask}}(\bm{Y}))) \nonumber \\
    \bm{\tilde{Y}} &= \mathsf{AddNorm}(\mathcal{F}(\bm{Z})),
\end{align}
where $\bm{Y}$ is the input sequence of vectors and $\selfatt_{\text{mask}}$ indicates a masked self-attention over time.
Finally, our decoder stacks together multiple decoder layers, helping to refine both the understanding of the textual input and the generation of next tokens.
Overall, the decoder takes as input word vectors, and the $t$-th element of its output sequence encodes the prediction of a word at time $t+1$, conditioned on $\bm{Y}_{\le t}$. After taking a linear projection and a softmax operation, this encodes a probability over words in the dictionary.

\subsection{Training details}
Following a standard practice in image captioning~\cite{ranzato2015sequence,rennie2017self,anderson2018bottom}, we pre-train our model with a word-level cross-entropy loss (XE) and finetune the sequence generation using reinforcement learning. When training with XE, the model is trained to predict the next token given previous ground-truth words; in this case, the input sequence for the decoder is immediately available and the computation of the entire output sequence can be done in a single pass, parallelizing all operations over time.

When training with reinforcement learning, we employ a variant of the self-critical sequence training approach~\cite{rennie2017self} on sequences sampled using beam search~\cite{anderson2018bottom}: to decode, we sample the top-$k$ words from the decoder probability distribution at each timestep, and always maintain the top-$k$ sequences with highest probability. As sequence decoding is iterative in this step, the aforementioned parallelism over time cannot be exploited. However, intermediate keys and values used to compute the output token at time $t$ can be reused in the next iterations.

Following previous works~\cite{anderson2018bottom}, we use the CIDEr-D score as reward, as it well correlates with human judgment~\cite{vedantam2015cider}. We baseline the reward using the mean of the rewards rather than greedy decoding as done in previous methods~\cite{rennie2017self,anderson2018bottom}, as we found it to slightly improve the final performance. 
The final gradient expression for one sample is thus:
\begin{equation}
    \nabla_\theta L(\theta) = -\frac{1}{k}\sum_{i=1}^k \left((r(\bm{w}^i)-b) \nabla_\theta \log p(\bm{w}^i)\right)
\end{equation}
where $\bm{w}^i$ is the $i$-th sentence in the beam, $r(\cdot)$ is the reward function, and $b = \left(\sum_i r(\bm{w}^i)\right)/k$ is the baseline, computed as the mean of the rewards obtained by the sampled sequences. At prediction time, we decode again using beam search, and keep the sequence with highest predicted probability among those in the last beam.

\section{Experiments}

\subsection{Datasets}
We first evaluate our model on the COCO dataset~\cite{lin2014microsoft}, which is the most commonly used test-bed for image captioning. Then, we assess the captioning of novel objects by testing on the recently proposed nocaps dataset~\cite{agrawal2019nocaps}.

\tinytit{COCO} The dataset contains more than $120\,000$ images, each of them annotated with $5$ different captions. We follow the splits provided by Karpathy~\etal~\cite{karpathy2015deep}, where $5\,000$ images are used for validation, $5\,000$ for testing and the rest for training. We also evaluate the model on the COCO online test server, composed of $40\,775$ images for which annotations are not made publicly available.

\tinytit{\texttt{\textbf{nocaps}}} The dataset consists of $15\,100$ images taken from the Open Images~\cite{kuznetsova2018open} validation and test sets, each annotated with $11$ human-generated captions. Images are divided into validation and test splits, respectively composed of $4\,500$ and $10\,600$ elements. Images can be further grouped into three subsets depending on the nearness to COCO, namely in-domain, near-domain, and out-of-domain images. Under this setting, we use COCO as training data and evaluate our results on the nocaps test server. 

\subsection{Experimental settings}
\tinytit{Metrics}
Following the standard evaluation protocol, we employ the full set of captioning metrics: BLEU~\cite{papineni2002bleu}, METEOR~\cite{banerjee2005meteor}, ROUGE~\cite{lin2004rouge},  CIDEr~\cite{vedantam2015cider}, and SPICE~\cite{spice2016}.

\tinytit{Implementation details} 
To represent image regions, we use Faster R-CNN~\cite{ren2015faster} with ResNet-101~\cite{he2016deep} finetuned on the Visual Genome dataset~\cite{krishnavisualgenome,anderson2018bottom}, thus obtaining a $2048$-dimensional feature vector for each region. To represent words, we use one-hot vectors and linearly project them to the input dimensionality of the model $d$. We also employ sinusoidal positional encodings~\cite{vaswani2017attention} to represent word positions inside the sequence and sum the two embeddings before the first decoding layer.

In our model, we set the dimensionality $d$ of each layer to $512$, the number of heads to $8$, and the number of memory vectors to $40$. We employ dropout with keep probability $0.9$ after each attention and feed-forward layer. In our meshed attention operator (Eq.~\ref{eq:mesh_attention}), we normalize the output with a scaling factor of $\sqrt{N}$.
Pre-training with XE is done following the learning rate scheduling strategy of~\cite{vaswani2017attention} with a warmup equal to $10\,000$ iterations. Then, during CIDEr-D optimization, we use a fixed learning rate of $5\times10^{-6}$. We train all models using the Adam optimizer~\cite{kingma2015adam}, a batch size of $50$, and a beam size equal to $5$. 

\tinytit{Novel object captioning} 
To train the model on the nocaps dataset, instead of using one-hot vectors, we represent words with GloVe word embeddings~\cite{pennington2014glove}. Two fully-connected layers are added to convert between the GloVe dimensionality and $d$ before the first decoding layer and after the last decoding layer. Before the final softmax, we multiply with the transpose of the word embeddings. All other implementation details are kept unchanged. 

\noindent Additional details on model architecture and training can be found in the supplementary material.

\begin{table}[t]
\small
\centering
\setlength{\tabcolsep}{.35em}
\resizebox{\linewidth}{!}{
\begin{tabular}{lcccccc}
\toprule
& B-1 & B-4 & M & R & C & S \\
\midrule
Transformer (w/ 6 layers as in~\cite{vaswani2017attention})  & 79.1 & 36.2 & 27.7 & 56.9 & 121.8 & 20.9 \\
Transformer (w/ 3 layers)   & 79.6 & 36.5 & 27.8 & 57.0 & 123.6 & 21.1 \\
Transformer (w/ AoA~\cite{huang2019attention})  & 80.3 & 38.8 & 29.0 & 58.4 & 129.1 & \textbf{22.7} \\
\midrule
$\mathcal{M}^2$ Transformer$^{\text{1-to-1}}$ (w/o mem.) & 80.5 & 38.2 & 28.9 & 58.2 & 128.4 & 22.2 \\
$\mathcal{M}^2$ Transformer$^{\text{1-to-1}}$ & 80.3 & 38.2 & 28.9 & 58.2 & 129.2 & 22.5 \\
\midrule
\ours (w/o mem.) & 80.4 & 38.3 & 29.0 & 58.2 & 129.4 & 22.6 \\
\ours (w/ softmax) & 80.3 & 38.4 & 29.1 & 58.3 & 130.3 & 22.5 \\
\textbf{\ours} & \textbf{80.8} & \textbf{39.1} & \textbf{29.2} & \textbf{58.6} & \textbf{131.2} & 22.6 \\

\bottomrule
\end{tabular}
}
\caption{Ablation study and comparison with Transformer-based alternatives. All results are reported after the REINFORCE optimization stage.}
\vspace{-.35cm}
\label{tab:ablation}
\end{table}

\subsection{Ablation study}
\tinytit{Performance of the Transformer}
In previous works, the Transformer model has been applied to captioning only in its original configuration with six layers, with the structure of connections that has been successful for uni-modal scenarios like machine translation. As we speculate that captioning requires specific architectures, we compare variations of the original Transformer with our approach.

Firstly, we investigate the impact of the number of encoding and decoding layers on captioning performance. As it can be seen in Table~\ref{tab:ablation}, the original Transformer (six layers) achieves $121.8$ CIDEr, slightly superior to the Up-Down approach~\cite{anderson2018bottom} which uses a two-layer recurrent language model with additive attention and includes a global feature vector ($120.1$ CIDEr). Varying the number of layers, we observe a significant increase in performance when using three encoding and three decoding layers, which leads to $123.6$ CIDEr. We hypothesize that this is due to the reduced training set size and to the lower semantic complexities of sentences in captioning with respect to those of language understanding tasks. Following this finding, all subsequent experiments will use three layers.

\tinytit{Attention on Attention baseline}
We also evaluate a recent proposal that can be straightforwardly applied to the Transformer as an alternative to standard dot-product attention. Specifically, we evaluate the addition of the ``Attention on Attention'' (AoA) approach~\cite{huang2019attention} to the attentive layers, both in the encoder and in the decoder. Noticeably, in~\cite{huang2019attention} this has been done with a Recurrent language model with attention, but the approach is sufficiently general to be applied to any attention stage. In this case, the result of dot-product attention is concatenated with the initial query and fed to two fully connected layers to obtain an information vector and a sigmoidal attention gate, then the two vectors are multiplied together. The final result is used as an alternative to the standard dot-product attention.
This addition to a standard Transformer with three layers leads to $129.1$ CIDEr (Table~\ref{tab:ablation}), thus underlying the usefulness of the approach also in Transformer-based models.

\begin{table}[t]
\small
\centering
\setlength{\tabcolsep}{.5em}
\resizebox{\linewidth}{!}{
\begin{tabular}{lccccccc}
\toprule
 & & B-1 & B-4 & M & R & C & S \\
\midrule
SCST~\cite{rennie2017self} & & - & 34.2 & 26.7 & 55.7 & 114.0 & - \\
Up-Down~\cite{anderson2018bottom} & & 79.8 & 36.3 & 27.7 & 56.9 & 120.1 & 21.4 \\
RFNet~\cite{jiang2018recurrent} & & 79.1 & 36.5 & 27.7 & 57.3 & 121.9 & 21.2 \\
Up-Down+HIP~\cite{yao2019hierarchy} & & - & 38.2 & 28.4 & 58.3 & 127.2 & 21.9 \\
GCN-LSTM~\cite{yao2018exploring} & & 80.5 & 38.2 & 28.5 & 58.3 & 127.6 & 22.0 \\
SGAE~\cite{yang2019auto} & & \textbf{80.8} & 38.4 & 28.4 & 58.6 & 127.8 & 22.1 \\ 
ORT~\cite{herdade2019image} & & 80.5 & 38.6 & 28.7 & 58.4 & 128.3 & \textbf{22.6} \\
AoANet~\cite{huang2019attention} & & 80.2 & 38.9 & \textbf{29.2} & \textbf{58.8} & 129.8 & 22.4 \\
\midrule
\textbf{\ours} & & \textbf{80.8} & \textbf{39.1} & \textbf{29.2} & 58.6 & \textbf{131.2} & \textbf{22.6} \\
\bottomrule
\end{tabular}
}
\caption{Comparison with the state of the art on the ``Karpathy'' test split, in single-model setting.}
\vspace{-.1cm}
\label{tab:sota_results}
\end{table}

\begin{table}[t]
\small
\centering
\setlength{\tabcolsep}{.5em}
\resizebox{\linewidth}{!}{
\begin{tabular}{lccccccc}
\toprule
 & & B-1 & B-4 & M & R & C & S \\
\midrule
\multicolumn{8}{c}{\textbf{Ensemble/Fusion of 2 models}} \\
\midrule
GCN-LSTM~\cite{yao2018exploring} & & 80.9 & 38.3 & 28.6 & 58.5 & 128.7 & 22.1 \\
SGAE~\cite{yang2019auto} & & 81.0 & 39.0 & 28.4 & 58.9 & 129.1 & 22.2 \\
ETA~\cite{liu2019entangled} & & 81.5 & \textbf{39.9} & 28.9 & 59.0 & 127.6 & 22.6 \\
GCN-LSTM+HIP~\cite{yao2019hierarchy} & & - & 39.1 & 28.9 & \textbf{59.2} & 130.6 & 22.3 \\
\midrule
\textbf{\ours} & & \textbf{81.6} & 39.8 & \textbf{29.5} & \textbf{59.2} & \textbf{133.2} & \textbf{23.1} \\
\midrule
\multicolumn{8}{c}{\textbf{Ensemble/Fusion of 4 models}} \\
\midrule
SCST~\cite{rennie2017self} & & - & 35.4 & 27.1 & 56.6 & 117.5 & - \\
RFNet~\cite{jiang2018recurrent} & & 80.4 & 37.9 & 28.3 & 58.3 & 125.7 & 21.7 \\
AoANet~\cite{huang2019attention} & & 81.6 & 40.2 & 29.3 & 59.4 & 132.0 & 22.8 \\
\midrule
\textbf{\ours} & & \textbf{82.0} & \textbf{40.5} & \textbf{29.7} & \textbf{59.5} & \textbf{134.5} & \textbf{23.5} \\
\bottomrule
\end{tabular}
}
\caption{Comparison with the state of the art on the  ``Karpathy'' test split, using an ensemble of models.}
\vspace{-.35cm}
\label{tab:sota_results_ensemble}
\end{table}

\begin{table*}[t]
\small
\centering
\setlength{\tabcolsep}{.54em}
\resizebox{\linewidth}{!}{
\begin{tabular}{lccccccccccccccccccccc}
\toprule
 & & \multicolumn{2}{c}{BLEU-1} & & \multicolumn{2}{c}{BLEU-2} & & \multicolumn{2}{c}{BLEU-3} & & \multicolumn{2}{c}{BLEU-4} & & \multicolumn{2}{c}{METEOR} & &  \multicolumn{2}{c}{ROUGE} & & \multicolumn{2}{c}{CIDEr} \\
\cmidrule{3-4} \cmidrule{6-7} \cmidrule{9-10} \cmidrule{12-13} \cmidrule{15-16} \cmidrule{18-19} \cmidrule{21-22} 
& & c5 & c40 & & c5 & c40 & & c5 & c40 & & c5 & c40 & & c5 & c40 & & c5 & c40 & & c5 & c40 \\
\midrule
SCST~\cite{rennie2017self} & & 78.1 & 93.7 & & 61.9 & 86.0 & & 47.0 & 75.9 & & 35.2 & 64.5 & & 27.0 & 35.5 & & 56.3 & 70.7 & & 114.7 & 116.7 \\
Up-Down~\cite{anderson2018bottom} & & 80.2 & 95.2 & & 64.1 & 88.8 & & 49.1 & 79.4 & & 36.9 & 68.5 & & 27.6 & 36.7 & & 57.1 & 72.4 & & 117.9 & 120.5 \\ 
RDN~\cite{ke2019reflective} & & 80.2 & 95.3 & & - & - & & - & - & & 37.3 & 69.5 & & 28.1 & 37.8 & & 57.4 & 73.3 & & 121.2 & 125.2 \\
RFNet~\cite{jiang2018recurrent} & & 80.4 & 95.0 & & 64.9 & 89.3 & & 50.1 & 80.1 & & 38.0 & 69.2 & & 28.2 & 37.2 & & 58.2 & 73.1 & & 122.9 & 125.1\\
GCN-LSTM~\cite{yao2018exploring} & & 80.8 & 95.9 & & 65.5 & 89.3 & & 50.8 & 80.3 & & 38.7 & 69.7 & & 28.5 & 37.6 & & 58.5 & 73.4 & & 125.3 & 126.5 \\
SGAE~\cite{yang2019auto} & & 81.0 & 95.3 & & 65.6 & 89.5 & & 50.7 & 80.4 & & 38.5 & 69.7 & & 28.2 & 37.2 & & 58.6 & 73.6 & & 123.8 & 126.5 \\
ETA~\cite{liu2019entangled} & & 81.2 & 95.0 & & 65.5 & 89.0 & & 50.9 & 80.4 & & 38.9 & 70.2 & & 28.6 & 38.0 & & 58.6 & 73.9 & & 122.1 & 124.4 \\
AoANet~\cite{huang2019attention} & & 81.0 & 95.0 & & 65.8 & 89.6 & & 51.4 & 81.3 & & 39.4 & 71.2 & & 29.1 & 38.5 & & 58.9 & 74.5 & & 126.9 & 129.6 \\
GCN-LSTM+HIP~\cite{yao2019hierarchy} & & \textbf{81.6} & 95.9 & & 66.2 & 90.4 & & 51.5 & 81.6 & & 39.3 & 71.0 & & 28.8 & 38.1 & & 59.0 & 74.1 & & 127.9 & 130.2 \\
\midrule
\textbf{\ours} & & \textbf{81.6} & \textbf{96.0} & & \textbf{66.4} & \textbf{90.8} & & \textbf{51.8} & \textbf{82.7} & & \textbf{39.7} & \textbf{72.8} & & \textbf{29.4} & \textbf{39.0} & & \textbf{59.2} & \textbf{74.8} & & \textbf{129.3} & \textbf{132.1} \\
\bottomrule
\end{tabular}
}
\caption{Leaderboard of various methods on the online MS-COCO test server.}
\vspace{-.35cm}
\label{tab:coco_test}
\end{table*}

\tinytit{Meshed Connectivity}
We then evaluate the role of the meshed connections between encoder and decoder layers. In Table~\ref{tab:ablation}, we firstly introduce a reduced version of our approach in which the $i$-th decoder layer is only connected to the corresponding $i$-th encoder layer (1-to-1), instead of being connected to all encoders. Using this 1-to-1 connectivity schema already brings an improvement with respect to using the output of the last encoder layer as in the standard Transformer ($123.6$ CIDEr vs $129.2$ CIDEr), thus confirming that exploiting a multi-level encoding of image regions is beneficial.
When we instead use our meshed connectivity schema, that exploits relationships encoded at all levels and weights them with a sigmoid gating, we observe a further performance improvement, from $129.2$ CIDEr to $131.2$ CIDEr. This amounts to a total improvement of $7.6$ points with respect to the standard Transformer. Also, the result of our full model is superior to that obtained using the AoA.

As an alternative to the sigmoid gating approach for weighting the contributions from different encoder layers (Eq.~\ref{eq:mesh_attention}), we also test with a softmax gating schema. In this case, the element-wise sigmoid applied to each encoder is replaced with a softmax operation over the rows of $\bm{\alpha}_i$. Using this alternative brings to a reduction of around 1 CIDEr point, underlying that it is beneficial to exploit the full potentiality of a weighted sum of the contributions from all encoding layers, rather than forcing a peaky distribution in which one layer is given more importance than the others.

\tinytit{Role of persistent memory}
We evaluate the role of memory vectors in both the 1-to-1 configuration and in the final configuration with meshed connections. As it can be seen from Table~\ref{tab:ablation}, removing memory vectors brings to a reduction in performance of around $1$ CIDEr point in both connectivity settings, thus confirming the usefulness of exploiting a priori learned knowledge when encoding image regions. Further experiments on the number of memory vectors can be found in the supplementary material.

\subsection{Comparison with state of the art}
We compare the performances of our approach with those of several recent proposals for image captioning.
The models we compare to include SCST~\cite{rennie2017self} and Up-Down~\cite{anderson2018bottom}, which respectively use attention over the grid of features and attention over regions. 
Also, we compare to RFNet~\cite{jiang2018recurrent}, which uses a recurrent fusion network to merge different CNN features; GCN-LSTM~\cite{yao2018exploring}, which exploits pairwise relationships between image regions through a Graph CNN; SGAE~\cite{yang2019auto}, which instead uses auto-encoding scene graphs. Further, we compare with the original AoANet~\cite{huang2019attention} approach, which uses attention on attention for encoding image regions and an LSTM language model. 
Finally, we compare with ORT~\cite{herdade2019image}, which uses a plain Transformer and weights attention scores in the region encoder with pairwise distances between detections. 

We evaluate our approach on the COCO ``Karpathy'' test split, using both single model and ensemble configurations, and on the online COCO evaluation server.

\begin{figure}[t]
    \centering
    \begin{minipage}{0.28\linewidth}
        \includegraphics[width=0.95\linewidth]{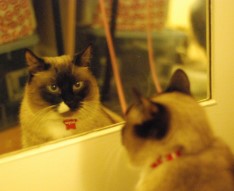}
        \end{minipage}
        \begin{minipage}{0.68\linewidth}
        \footnotesize{
        \textbf{GT:} A cat looking at his reflection in the mirror.\\
        \textbf{Transformer:} A cat sitting in a window sill looking out.\\
        \textbf{\ours:} A cat looking at its reflection in a mirror.
        }
    \end{minipage}

    \vspace{0.1cm}
    
    \begin{minipage}{0.28\linewidth}
        \includegraphics[width=0.95\linewidth]{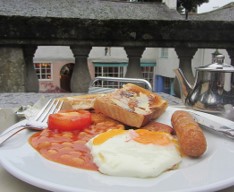}
        \end{minipage}
        \begin{minipage}{0.68\linewidth}
        \footnotesize{
        \textbf{GT:} A plate of food including eggs and toast on a table next to a stone railing.\\
        \textbf{Transformer:} A group of food on a plate.\\
        \textbf{\ours:} A plate of breakfast food with eggs and toast.
        }
    \end{minipage}

    \vspace{0.1cm}
    
    \begin{minipage}{0.28\linewidth}
        \includegraphics[width=0.95\linewidth]{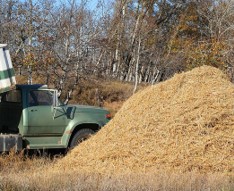}
        \end{minipage}
        \begin{minipage}{0.68\linewidth}
        \footnotesize{
        \textbf{GT:} A truck parked near a tall pile of hay.\\
        \textbf{Transformer:} A truck is parked in the grass in a field.\\
        \textbf{\ours:} A green truck parked next to a pile of hay.
        }
    \end{minipage}

    \caption{Examples of captions generated by our approach and the original Transformer model, as well as the corresponding ground-truths.}
    \label{fig:qualitative_results}
\vspace{-.35cm}
\end{figure}

\begin{figure*}
    \centering
    \setlength{\tabcolsep}{.025em}
    \renewcommand*{\arraystretch}{0.5}
    \begin{tabular}{cccccccc}
    \includegraphics[width=0.124\linewidth]{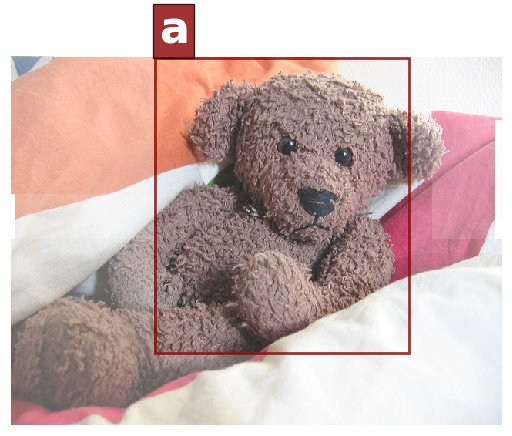} & 
    \includegraphics[width=0.124\linewidth]{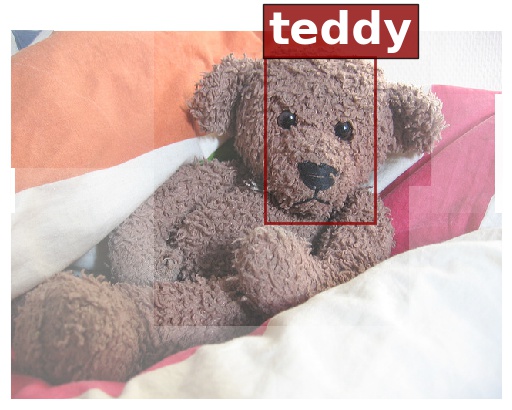} & 
    \includegraphics[width=0.124\linewidth]{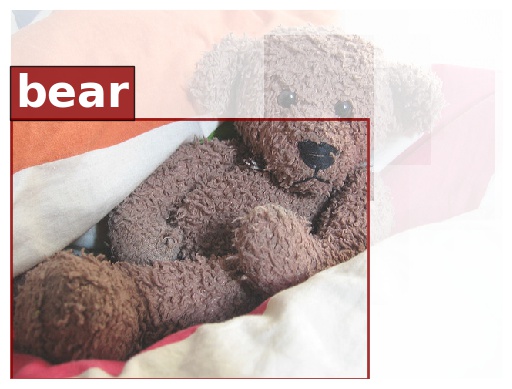} & 
    \includegraphics[width=0.124\linewidth]{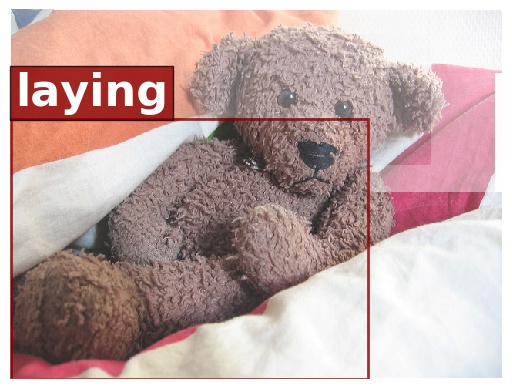} & 
    \includegraphics[width=0.124\linewidth]{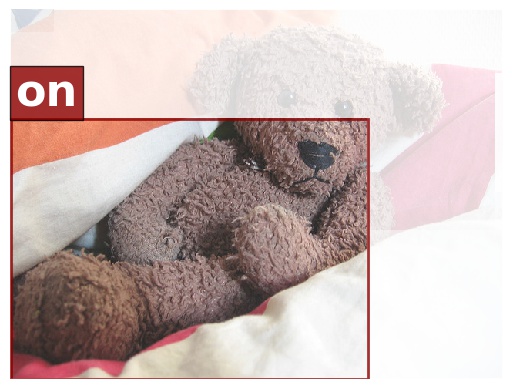} & 
    \includegraphics[width=0.124\linewidth]{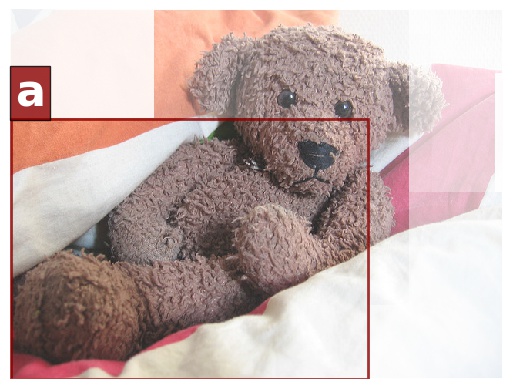} &
    \includegraphics[width=0.124\linewidth]{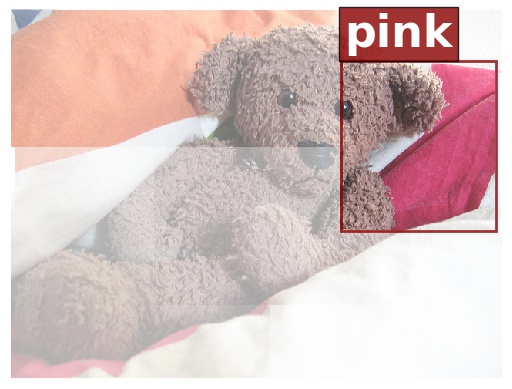} & 
    \includegraphics[width=0.124\linewidth]{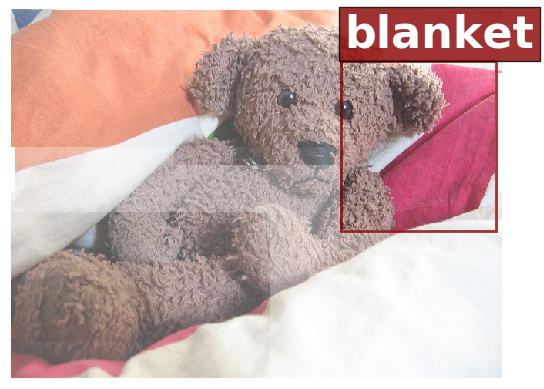}\\
    \includegraphics[width=0.124\linewidth]{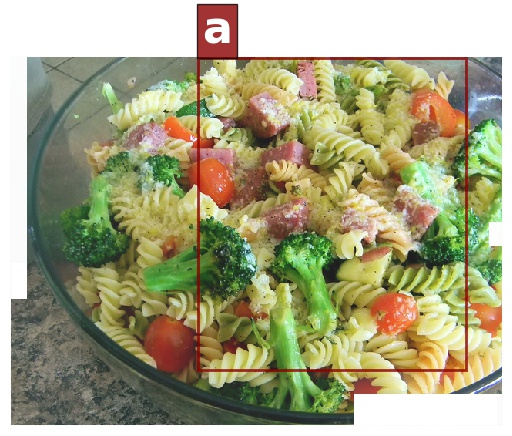} & 
    \includegraphics[width=0.124\linewidth]{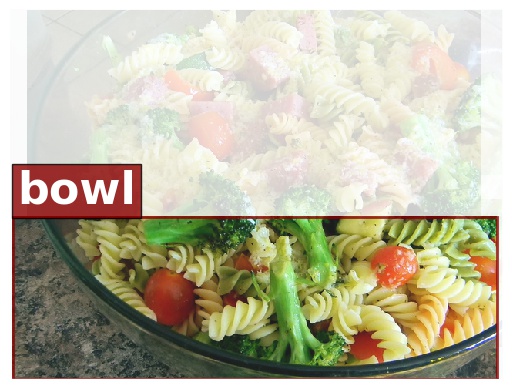} & 
    \includegraphics[width=0.124\linewidth]{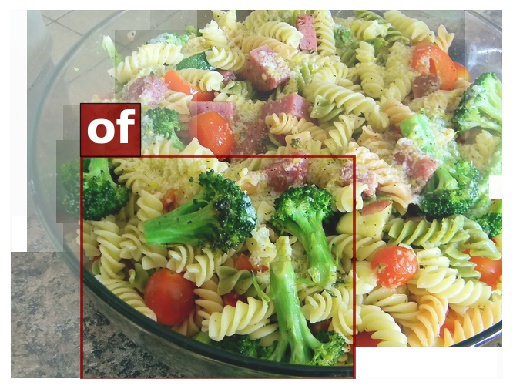} & 
    \includegraphics[width=0.124\linewidth]{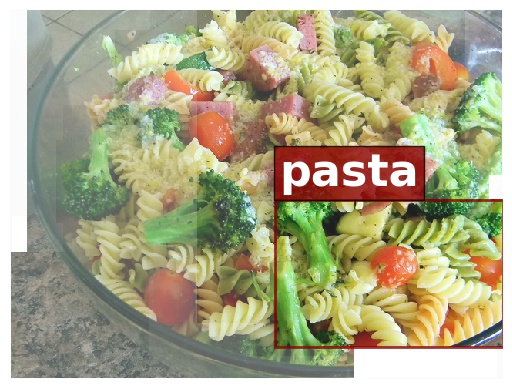} & 
    \includegraphics[width=0.124\linewidth]{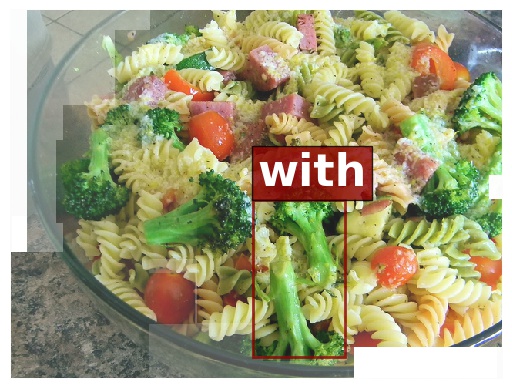} & 
    \includegraphics[width=0.124\linewidth]{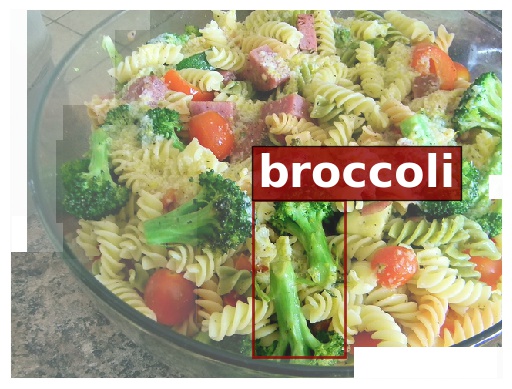} &
    \includegraphics[width=0.124\linewidth]{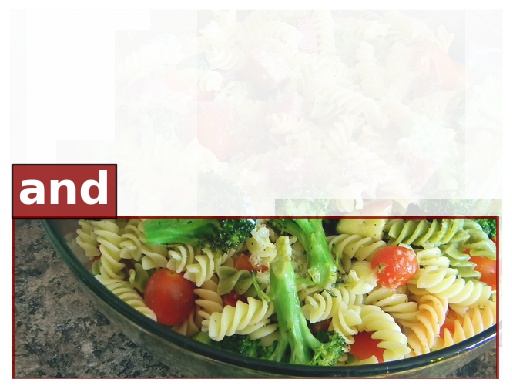} & 
    \includegraphics[width=0.124\linewidth]{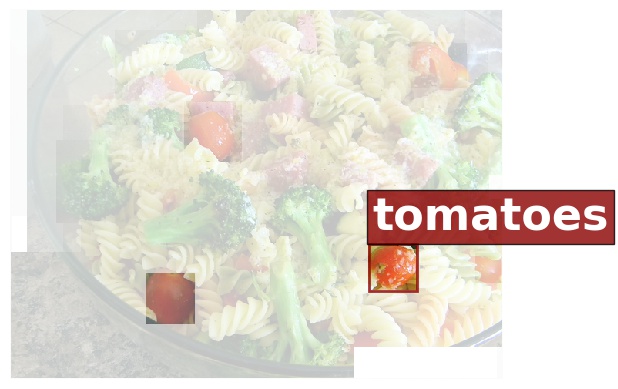} \\
    \includegraphics[width=0.124\linewidth]{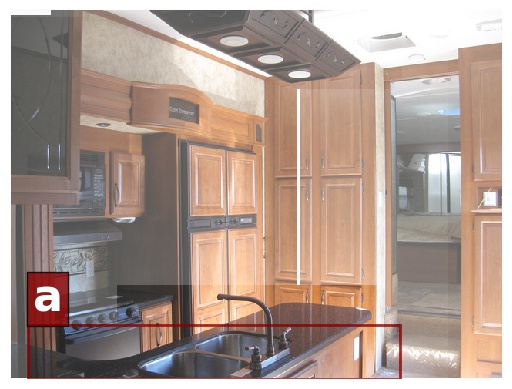} & 
    \includegraphics[width=0.124\linewidth]{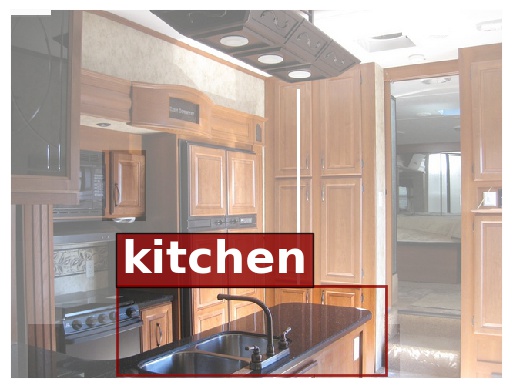} & 
    \includegraphics[width=0.124\linewidth]{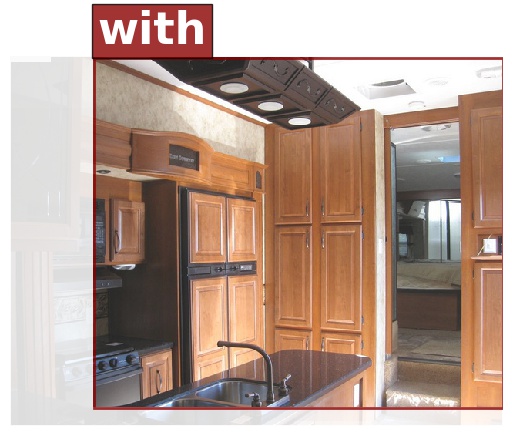} & 
    \includegraphics[width=0.124\linewidth]{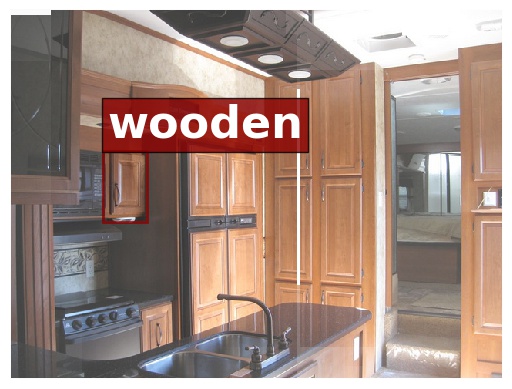} & 
    \includegraphics[width=0.124\linewidth]{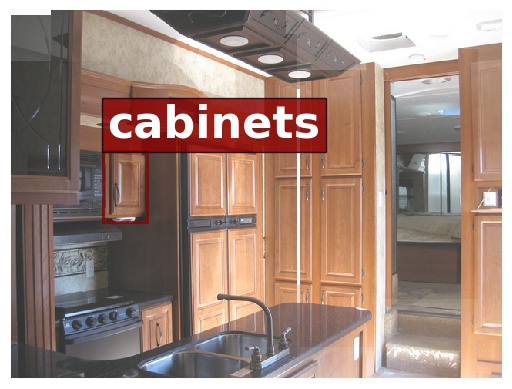} & 
    \includegraphics[width=0.124\linewidth]{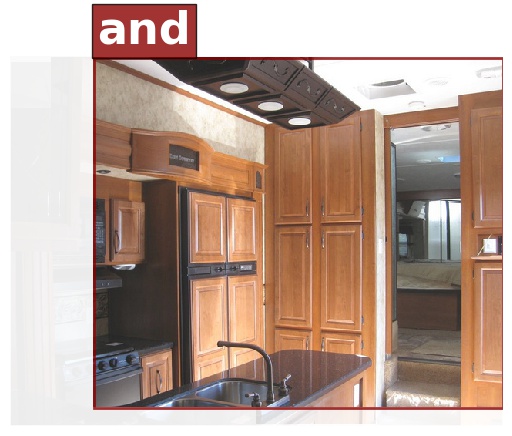} &
    \includegraphics[width=0.124\linewidth]{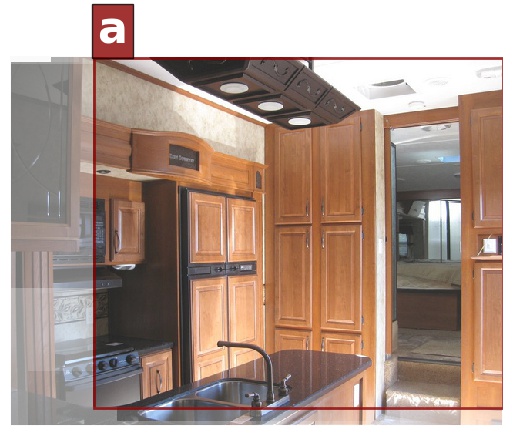} & 
    \includegraphics[width=0.124\linewidth]{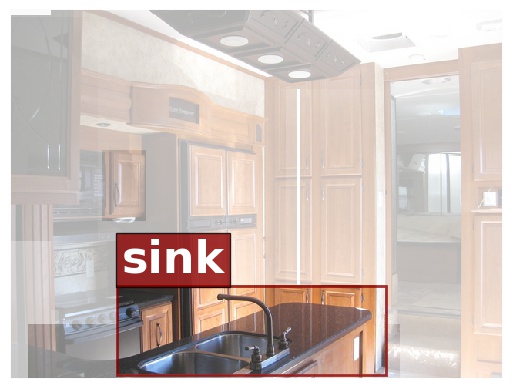}\\
    \end{tabular}
    \caption{Visualization of attention states for three sample captions. For each generated word, we show the attended image regions, outlining the region with the maximum output attribution in red.}
    \label{fig:attention}
\vspace{-.1cm}
\end{figure*}

\tinytit{Single model}
In Table~\ref{tab:sota_results} we report the performance of our method in comparison with the aforementioned competitors, using captions predicted from a single model and optimization on the CIDEr-D score. As it can be observed, our method surpasses all other approaches in terms of BLEU-4, METEOR and CIDEr, while being competitive on BLEU-1 and SPICE with the best performer, and slightly worse on ROUGE with respect to AoANet~\cite{huang2019attention}. In particular, it advances the current state of the art on CIDEr by 1.4 points.

\tinytit{Ensemble model}
Following the common practice~\cite{rennie2017self,huang2019attention} of building an ensemble of models, we also report the performances of our approach when averaging the output probability distributions of multiple and independently trained instances of our model. In Table~\ref{tab:sota_results_ensemble}, we use ensembles of two and four models, trained from different random seeds. Noticeably, when using four models our approach achieves the best performance according to all metrics, with an increase of 2.5 CIDEr points with respect to the current state of the art~\cite{huang2019attention}.

\tinytit{Online Evaluation}
Finally, we also report the performance of our method on the online COCO test server\footnote{\scriptsize\url{https://competitions.codalab.org/competitions/3221}}. In this case, we use the ensemble of four models previously described, trained on the ``Karpathy'' training split. The evaluation is done on the COCO test split, for which ground-truth annotations are not publicly available. Results are reported in Table~\ref{tab:coco_test}, in comparison with the top-performing approaches of the leaderboard. For fairness of comparison, they also used an ensemble configuration. As it can be seen, our method surpasses the current state of the art on all metrics, achieving an advancement of 1.4 CIDEr points with respect to the best performer.

\subsection{Describing novel objects}
We also assess the performance of our approach when dealing with images containing object categories that are not seen in the training set. We compare with Up-Down~\cite{anderson2018bottom} and Neural Baby Talk~\cite{lu2018neural}, when using GloVe word embeddings and Constrained Beam Search (CBS)~\cite{anderson2016guided} to address the generation of out-of-vocabulary words and constrain the presence of categories detected by an object detector. To compare with our model, we use a simplified implementation of the procedure described in~\cite{agrawal2019nocaps} to extract constraints, without using word phrases. 
Results are shown in Table~\ref{tab:nocaps}: as it can be seen, the original Transformer is significantly less performing than Up-Down on both in-domain and out-of-domain categories, while our approach can properly deal with novel categories, surpassing the Up-Down baseline in both in-domain and out-of-domain images. As expected, the use of CBS significantly enhances the performances, in particular on out-of-domain captioning.

\begin{table}[t]
\small
\centering
\setlength{\tabcolsep}{.35em}
\resizebox{\linewidth}{!}{
\begin{tabular}{lccccccccc}
\toprule
 & & \multicolumn{2}{c}{In-Domain} & & \multicolumn{2}{c}{Out-of-Domain} & & \multicolumn{2}{c}{Overall} \\
\cmidrule{3-4} \cmidrule{6-7} \cmidrule{9-10}
& & CIDEr & SPICE & & CIDEr & SPICE & & CIDEr & SPICE \\
\midrule
NBT~+~CBS~\cite{agrawal2019nocaps}      & & 62.1 & 10.1 & & 62.4 & 8.9 & & 60.2 & 9.5 \\
Up-Down~+~CBS~\cite{agrawal2019nocaps}  & & 80.0 & 12.0 & & 66.4 & 9.7 & & 73.1 & 11.1 \\
\midrule
Transformer                       & & 78.0 & 11.0 & & 29.7 & 7.8 & & 54.7 & 9.8 \\
\textbf{\ours}                    & & \textbf{85.7} & \textbf{12.1} & & 38.9 & 8.9 & & 64.5 & 11.1 \\
\midrule
Transformer~+~CBS                 & & 74.3 & 11.0 & & 62.5 & 9.2 & & 66.9 & 10.3 \\
\textbf{\ours+~CBS}               & & 81.2 & 12.0 & & \textbf{69.4} & \textbf{10.0} & & \textbf{75.0} & \textbf{11.4} \\
\bottomrule
\end{tabular}
}
\caption{Performances on nocaps validation set, for in-domain and out-of-domain captioning.}
\label{tab:nocaps}
\vspace{-.2cm}
\end{table}

\subsection{Qualitative results and visualization}
Figure~\ref{fig:qualitative_results} proposes qualitative results generated by our model and the original Transformer. On average, our model is able to generate more accurate and descriptive captions, integrating fine-grained details and object relations. 

Finally, to better understand the effectiveness of our \ours, we investigate the contribution of detected regions to the model output. Differently from recurrent-based captioning models, in which attention weights over regions can be easily extracted, in our model the contribution of one region with respect to the output is given by more complex non-linear dependencies. Therefore, we revert to attribution methods: specifically, we employ the Integrated Gradients approach~\cite{sundararajan2017axiomatic}, which approximates the integral of gradients with respect to the given input. Results are presented in Figure~\ref{fig:attention}, where we observe that our approach correctly grounds image regions to words, also in presence of object details and small detections.
More visualizations are included in the supplementary material.

\section{Conclusion}
We presented \ours, a novel Transformer-based architecture for image captioning.
Our model incorporates a region encoding approach that exploits a priori knowledge through memory vectors and a meshed connectivity between encoding and decoding modules. Noticeably, this connectivity pattern is unprecedented for other fully-attentive architectures. Experimental results demonstrated that our approach achieves a new state of the art on COCO, ranking first in the on-line leaderboard. Finally, we validated the components of our model through ablation studies, and its performances when describing novel objects.

\section*{Acknowledgment}
\small{
This work was partially supported by the ``IDEHA - Innovation for Data Elaboration in Heritage Areas'' project (PON ARS01\_00421), funded by the Italian Ministry of Education (MIUR). We also acknowledge the NVIDIA AI Technology Center, EMEA, for its support and access to computing resources.}

{\small
\bibliographystyle{ieee_fullname}
\bibliography{egbib}
}

\newpage
\appendix

\section{Supplementary material}
In the following, we present additional material about our \ours model. In particular, we provide additional training and implementation details, further experimental results, and visualizations.

\subsection{Additional implementation details}
\tinytit{Decoding optimization}
As mentioned in Sec.~3.3, during the decoding stage computation cannot be parallelized over time as the input sequence is iteratively built. A naive approach would be to feed the model at each iteration with the previous $t-1$ generated words, $\{w_0, w_1, ..., w_{t-1}\}$ and sample the next predicted word $w_t$ after computing the results of each attention and feed-forward layer over all timesteps. This in practice requires to re-compute the same queries, keys, values and attentive states multiple times, with intermediate results depending on $w_t$ being recomputed $T-t$ times, where $T$ is the length of the sampled sequence (in our experiments $T$ is equal to $20$).

In our implementation, we revert to a more computationally friendly approach in which we re-use intermediate results computed at previous timesteps. Each attentive layer of the decoder internally stores previously computed keys and values. At each timestep of the decoding, the model is fed only with $w_{t-1}$, and we only compute queries, keys and values depending on $w_{t-1}$. 

In PyTorch, this can be implemented by exploiting the \texttt{register\_buffer} method of \texttt{nn.Module}, and creating buffers to hold previously computed results. When running on a NVIDIA 2080Ti GPU, we found this to reduce training and inference times by approximately a factor of 3.

\tinytit{Vocabulary and tokenization}
We convert all captions to lowercase, remove punctuation characters and tokenize using the spaCy NLP toolkit\footnote{\url{https://spacy.io/}}. To build vocabularies, we remove all words which appear less than $5$ times in training and validation splits.
For each image, we use a maximum number of region feature vectors equal to $50$. 

\tinytit{Model dimensionality and weight initialization}
Using $8$ attentive heads, the size of queries, keys and values in each head is set to $d/8 = 64$.
Weights of attentive layers are initialized from the uniform distribution proposed by Glorot \etal~\cite{glorot2010understanding}, while weights of feed-forward layers are initialized using~\cite{he2015delving}. All biases are initialized to 0. Memory vectors for keys and values are initialized from a normal distribution with zero mean and, respectively, $1/d_k$ and $1/m$ variance, where $d_k$ is the dimensionality of keys and $m$ is the number of memory vectors.

\begin{table}[t]
\small
\centering
\setlength{\tabcolsep}{.5em}
\begin{tabular}{cccccccc}
\toprule
Memories & & B-1 & B-4 & M & R & C & S\\
\midrule
No memory & & 80.4 & 38.3 & 29.0 & 58.2 & 129.4 & 22.6 \\
20 & & 80.7 & 38.9 & 29.0 & 58.4 & 129.9 & 22.7 \\
\textbf{40} & & \textbf{80.8} & \textbf{39.1} & \textbf{29.2} & \textbf{58.6} & \textbf{131.2} & 22.6 \\
60 & & 80.0 & 37.9 & 28.9 & 58.1 & 129.6 & 22.5 \\
80 & & 80.0 & 38.2 & 29.0 & 58.3 & 128.9 & \textbf{22.9} \\
\bottomrule
\end{tabular}
\caption{Captioning results of \ours using different numbers of memory vectors.}
\label{tab:memory}
\end{table}

\begin{table}[t]
\small
\centering
\setlength{\tabcolsep}{.5em}
\begin{tabular}{cccccccc}
\toprule
Layers & & B-1 & B-4 & M & R & C & S\\
\midrule
2 & & 80.5 & 38.6 & 29.0 & 58.4 & 128.5 & \textbf{22.8} \\
3 & & \textbf{80.8} & \textbf{39.1} & \textbf{29.2} & \textbf{58.6} & \textbf{131.2} & 22.6 \\
4 & & \textbf{80.8} & 38.6 & 29.1 & 58.5 & 129.6 & 22.6 \\
\bottomrule
\end{tabular}
\caption{Captioning results of \ours using different numbers of encoder and decoder layers.}
\label{tab:layers}
\vspace{-0.3cm}
\end{table}

\subsection{Additional experimental results}

\tinytit{Memory vectors}
In Table~\ref{tab:memory}, we report the performance of our approach when using a varying number of memory vectors. As it can be seen, the best result in terms of BLEU, METEOR, ROUGE and CIDEr is obtained with $40$ memory vectors, while $80$ memory vectors provide a slightly superior result in terms of SPICE. Therefore, all experiments in the main paper are carried out with $40$ memory vectors.

\begin{table}[t]
\small
\centering
\setlength{\tabcolsep}{.35em}
\resizebox{\linewidth}{!}{
\begin{tabular}{lccccccc}
\toprule
& SPICE & Obj. & Attr. & Rel. & Color & Count & Size \\
\midrule
Up-Down~\cite{anderson2018bottom} & 21.4 & 39.1 & 10.0 & 6.5 & 11.4 & 18.4 & 3.2 \\
\midrule
Transformer & 21.1 & 38.6 & 9.6 & 6.3 & 9.2 & 17.5 & 2.0 \\
\textbf{\ours} & \textbf{22.6} & \textbf{40.0} & \textbf{11.6} & \textbf{6.9} & \textbf{12.9} & \textbf{20.4} & \textbf{3.5} \\
\bottomrule
\end{tabular}
}
\caption{Breakdown of SPICE F-scores over various subcategories.}
\vspace{-.3cm}
\label{tab:spice}
\end{table}

\tit{Encoder and decoder layers}
To complement the analysis presented in Sec.~4.3, we also investigate the performance of the \ours when changing the number of encoding and decoding layers. Table~\ref{tab:layers} shows that the best performance is obtained with three encoding and decoding layers, thus confirming the initial findings on the base Transformer model. As our model can deal with a different number of encoding and decoding layers, we also experimented with non symmetric encoding-decoding architectures, without however noticing significant improvements in performance.

\tit{SPICE F-scores}
Finally, in Table~\ref{tab:spice} we report a breakdown of SPICE F-scores over various subcategories on the ``Karpathy'' test split, in comparison with the Up-Down approach~\cite{anderson2018bottom} and the base Transformer model with three layers. As it can be seen, our model significantly improves on identifying objects, attributes and relationships between objects.

\begin{figure*}[t]
    \centering
    \begin{minipage}{0.16\linewidth}
        \includegraphics[width=0.95\linewidth]{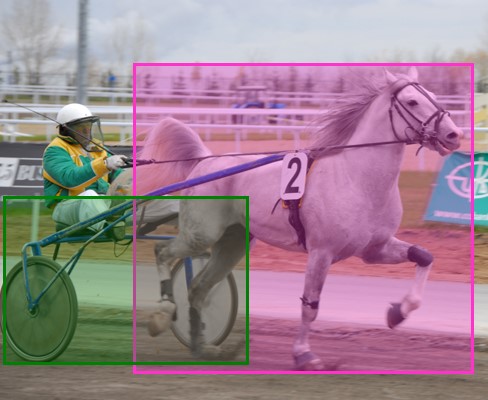}
        \end{minipage}
    \begin{minipage}{0.3\linewidth}
        \footnotesize{
        \textbf{Constraints:} \textcolor{magenta}{horse}; \textcolor{darkgreen}{cart}.\\
        \vspace{0.03cm}
         
        \textbf{Transformer:} A \textcolor{magenta}{horse} pulling a \textcolor{darkgreen}{cart} down a street.\\
        \textbf{\ours:} A white \textcolor{magenta}{horse} pulling a man in a \textcolor{darkgreen}{cart}.
        }
    \end{minipage}
    \hspace{0.2cm}
    \begin{minipage}{0.16\linewidth}
        \includegraphics[width=0.95\linewidth]{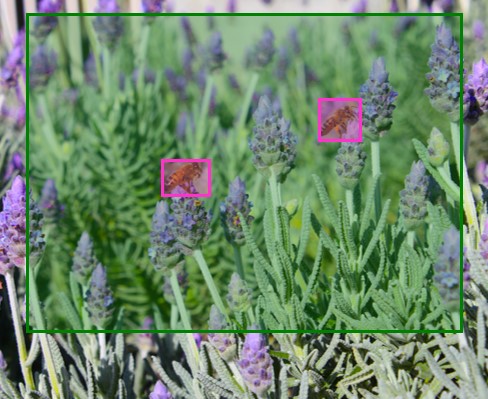}
        \end{minipage}
    \begin{minipage}{0.3\linewidth}
        \footnotesize{
        \textbf{Constraints:} \textcolor{magenta}{bee}; \textcolor{darkgreen}{lavender}.\\
        \vspace{0.03cm}
         
        \textbf{Transformer:} A \textcolor{magenta}{bee} \textcolor{darkgreen}{lavender} of purple flowers in a field.\\
        \textbf{\ours:} A field of \textcolor{darkgreen}{lavender} purple flowers with \textcolor{magenta}{bee}.
        }
    \end{minipage}

    \vspace{0.15cm}

    \begin{minipage}{0.16\linewidth}
        \includegraphics[width=0.95\linewidth]{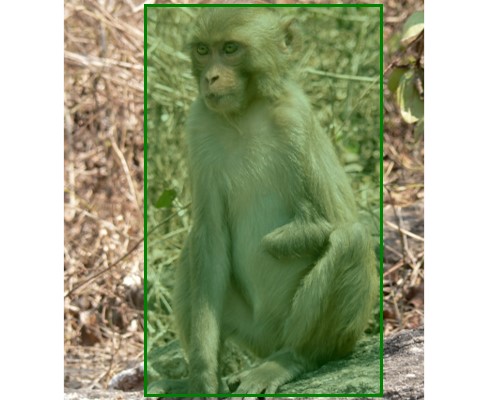}
        \end{minipage}
    \begin{minipage}{0.3\linewidth}
        \footnotesize{
        \textbf{Constraints:} \textcolor{darkgreen}{monkey}.\\
        \vspace{0.03cm}
         
        \textbf{Transformer:} A brown bear sitting on a rock \textcolor{darkgreen}{monkey}.\\
        \textbf{\ours:} A small \textcolor{darkgreen}{monkey} sitting on a rock in the grass.
        }
    \end{minipage}
    \hspace{0.2cm}
    \begin{minipage}{0.16\linewidth}
        \includegraphics[width=0.95\linewidth]{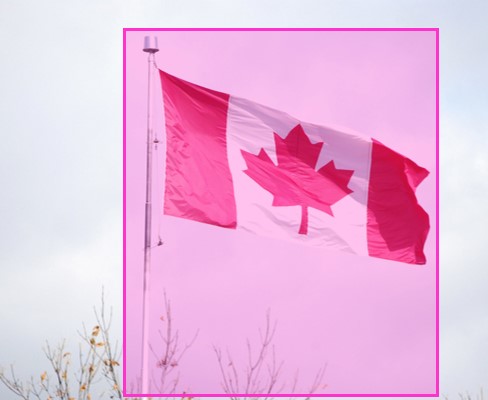}
        \end{minipage}
        \begin{minipage}{0.3\linewidth}
        \footnotesize{
        \textbf{Constraints:} \textcolor{magenta}{flag}.\\
        \vspace{0.03cm}
         
        \textbf{Transformer:} A red kite with a \textcolor{magenta}{flag} in the sky.\\
        \textbf{\ours:} A red and white \textcolor{magenta}{flag} flying in the sky.
        }
    \end{minipage}
                
    \vspace{0.15cm}

    \begin{minipage}{0.16\linewidth}
        \includegraphics[width=0.95\linewidth]{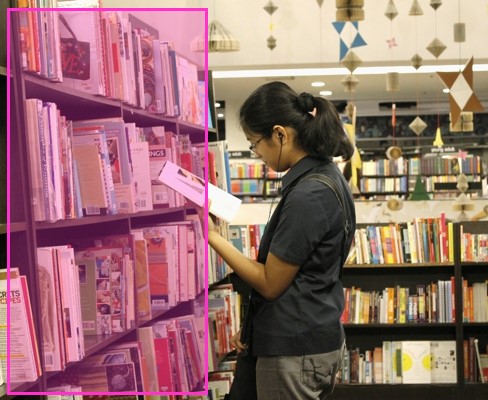}
        \end{minipage}
    \begin{minipage}{0.3\linewidth}
        \footnotesize{
        \textbf{Constraints:} \textcolor{magenta}{bookcase}.\\
        \vspace{0.03cm}
         
        \textbf{Transformer:} A woman holding a \textcolor{magenta}{bookcase} in a store.\\
        \textbf{\ours:} A woman holding a book in front of a \textcolor{magenta}{bookcase}.
        }
    \end{minipage}
    \hspace{0.2cm}
    \begin{minipage}{0.16\linewidth}
        \includegraphics[width=0.95\linewidth]{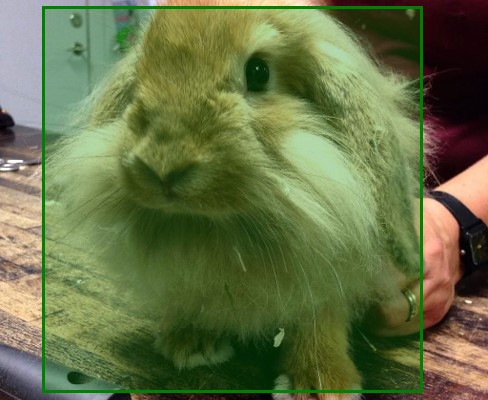}
        \end{minipage}
        \begin{minipage}{0.3\linewidth}
        \footnotesize{
        \textbf{Constraints:} \textcolor{darkgreen}{rabbit}.\\
        \vspace{0.03cm}
         
        \textbf{Transformer:} A cat sitting on the \textcolor{darkgreen}{rabbit} with a cell phone.\\
        \textbf{\ours:} A \textcolor{darkgreen}{rabbit} sitting on a table next to a person.
        }
    \end{minipage}
\caption{Sample nocaps images and corresponding predicted captions generated by our model and the original Transformer. For each image, we report the Open Images object classes predicted by the object detector and used as constraints during the generation of the caption.}
\label{fig:nocaps}
\vspace{-.3cm}
\end{figure*}

\subsection{Qualitative results and visualization}
\begin{figure*}[t]
    \centering
    \begin{minipage}{0.16\linewidth}
        \includegraphics[width=0.95\linewidth]{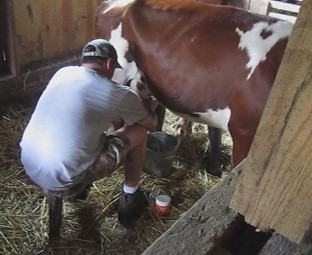}
        \end{minipage}
    \begin{minipage}{0.3\linewidth}
        \footnotesize{
        \textbf{GT:} A man milking a brown and white cow in barn.\\
        \textbf{Transformer:} A man is standing next to a cow.\\
        \textbf{\ours:} A man is milking a cow in a barn.
        }
    \end{minipage}
    \hspace{0.2cm}
    \begin{minipage}{0.16\linewidth}
        \includegraphics[width=0.95\linewidth]{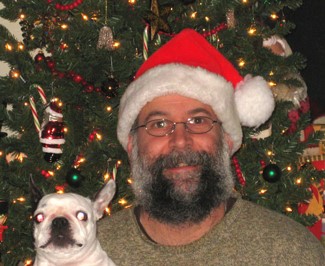}
        \end{minipage}
        \begin{minipage}{0.3\linewidth}
        \footnotesize{
        \textbf{GT:} A man in a red Santa hat and a dog pose in front of a Christmas tree.\\
        \textbf{Transformer:} A Christmas tree in the snow with a Christmas tree.\\
        \textbf{\ours:} A man wearing a Santa hat with a dog in front of a Christmas tree.
        }
    \end{minipage}
                
    \vspace{0.15cm}
    
    \begin{minipage}{0.16\linewidth}
        \includegraphics[width=0.95\linewidth]{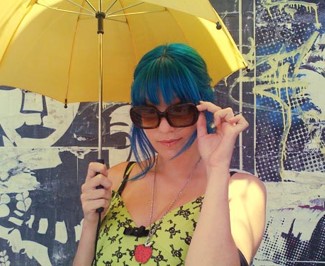}
        \end{minipage}
    \begin{minipage}{0.3\linewidth}
        \footnotesize{
        \textbf{GT:} A woman with blue hair and a yellow umbrella.\\
        \textbf{Transformer:} A woman is holding an umbrella.\\
        \textbf{\ours:} A woman with blue hair holding a yellow umbrella.
        }
    \end{minipage}
    \hspace{0.2cm}
    \begin{minipage}{0.16\linewidth}
        \includegraphics[width=0.95\linewidth]{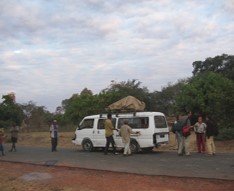}
        \end{minipage}
    \begin{minipage}{0.3\linewidth}
        \footnotesize{
        \textbf{GT:} Several people standing outside a parked white van.\\
        \textbf{Transformer:} A group of people standing outside of a bus.\\
        \textbf{\ours:} A group of people standing around a white van.
        }
    \end{minipage}
                    
    \vspace{0.15cm}
    
    \begin{minipage}{0.16\linewidth}
        \includegraphics[width=0.95\linewidth]{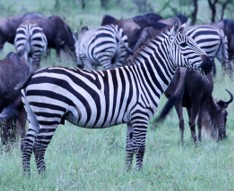}
        \end{minipage}
    \begin{minipage}{0.3\linewidth}
        \footnotesize{
        \textbf{GT:} Several zebras and other animals grazing in a field.\\
        \textbf{Transformer:} A herd of zebras are standing in a field.\\
        \textbf{\ours:} A herd of zebras and other animals grazing in a field.
        }
    \end{minipage}
    \hspace{0.2cm}
    \begin{minipage}{0.16\linewidth}
        \includegraphics[width=0.95\linewidth]{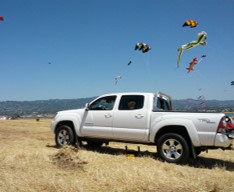}
        \end{minipage}
    \begin{minipage}{0.3\linewidth}
        \footnotesize{
        \textbf{GT:} A truck sitting on a field with kites in the air.\\
        \textbf{Transformer:} A group of cars parked in a field with a kite.\\
        \textbf{\ours:} A white truck is parked in a field with kites.
        }
    \end{minipage}
                    
    \vspace{0.15cm}
    
    \begin{minipage}{0.16\linewidth}
        \includegraphics[width=0.95\linewidth]{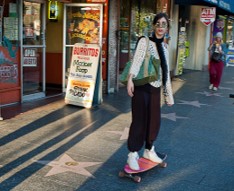}
        \end{minipage}
    \begin{minipage}{0.3\linewidth}
        \footnotesize{
        \textbf{GT:} A woman who is skateboarding down the street.\\
        \textbf{Transformer:} A woman walking down a street talking on a cell phone. \\
        \textbf{\ours:} A woman standing on a skateboard on a street.
        }
    \end{minipage}
    \hspace{0.2cm}
    \begin{minipage}{0.16\linewidth}
        \includegraphics[width=0.95\linewidth]{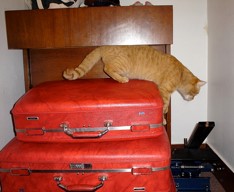}
        \end{minipage}
    \begin{minipage}{0.3\linewidth}
        \footnotesize{
        \textbf{GT:} Orange cat walking across two red suitcases stacked on floor.\\
        \textbf{Transformer:} An orange cat sitting on top of a suitcase.\\
        \textbf{\ours:} An orange cat standing on top of two red suitcases.
        }
    \end{minipage}
                    
    \vspace{0.15cm}
    
    \begin{minipage}{0.16\linewidth}
        \includegraphics[width=0.95\linewidth]{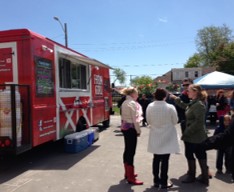}
        \end{minipage}
    \begin{minipage}{0.3\linewidth}
        \footnotesize{
        \textbf{GT:} Some people are standing in front of a red food truck.\\
        \textbf{Transformer:} A group of people standing in front of a bus.\\
        \textbf{\ours:} A group of people standing outside of a food truck.
        }
    \end{minipage}
    \hspace{0.2cm}
    \begin{minipage}{0.16\linewidth}
        \includegraphics[width=0.95\linewidth]{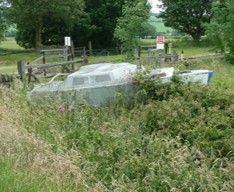}
        \end{minipage}
    \begin{minipage}{0.3\linewidth}
        \footnotesize{
        \textbf{GT:} A boat parked in a field with long green grass.\\
        \textbf{Transformer:} A field of grass with a fence.\\
        \textbf{\ours:} A boat in the middle of a field of grass.
        }
    \end{minipage}
                    
    \vspace{0.15cm}
    
    \begin{minipage}{0.16\linewidth}
        \includegraphics[width=0.95\linewidth]{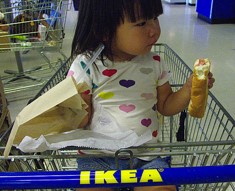}
        \end{minipage}
    \begin{minipage}{0.3\linewidth}
        \footnotesize{
        \textbf{GT:} A little girl is eating a hot dog and riding in a shopping cart.\\
        \textbf{Transformer:} A little girl sitting on a bench eating a hot dog.\\
        \textbf{\ours:} A little girl sitting in a shopping cart eating a hot dog.
        }
    \end{minipage}
    \hspace{0.2cm}
    \begin{minipage}{0.16\linewidth}
        \includegraphics[width=0.95\linewidth]{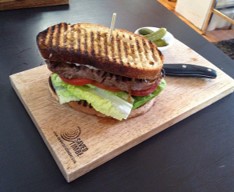}
        \end{minipage}
    \begin{minipage}{0.3\linewidth}
        \footnotesize{
        \textbf{GT:} A grilled sandwich sits on a cutting board by a knife.\\
        \textbf{Transformer:} A sandwich sitting on top of a wooden table.\\
        \textbf{\ours:} A sandwich on a cutting board with a knife.
        }
    \end{minipage}
                        
    \vspace{0.15cm}
    
    \begin{minipage}{0.16\linewidth}
        \includegraphics[width=0.95\linewidth]{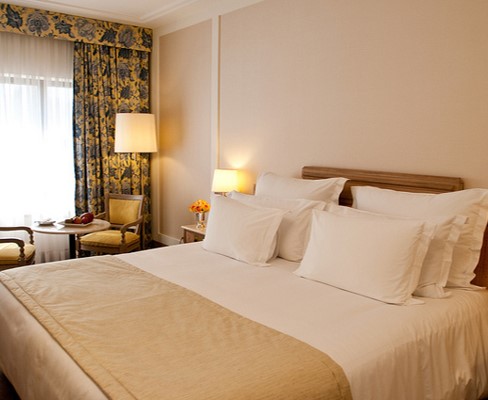}
        \end{minipage}
    \begin{minipage}{0.3\linewidth}
        \footnotesize{
        \textbf{GT:} A hotel room with a well-made bed, a table, and two chairs.\\
        \textbf{Transformer:} A bedroom with a bed and a table.\\
        \textbf{\ours:} A hotel room with a large bed with white pillows.
        }
    \end{minipage}
    \hspace{0.2cm}
    \begin{minipage}{0.16\linewidth}
        \includegraphics[width=0.95\linewidth]{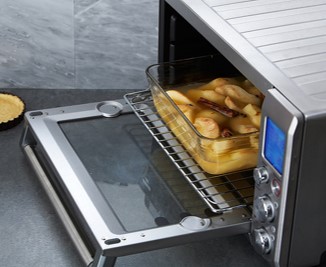}
        \end{minipage}
    \begin{minipage}{0.3\linewidth}
        \footnotesize{
        \textbf{GT:} An open toaster oven with a glass dish of food inside.\\
        \textbf{Transformer:} An open suitcase with food in an oven.\\
        \textbf{\ours:} A toaster oven with a tray of food inside of it.
        }
    \end{minipage}
                        
    \vspace{0.15cm}
    
    \begin{minipage}{0.16\linewidth}
        \includegraphics[width=0.95\linewidth]{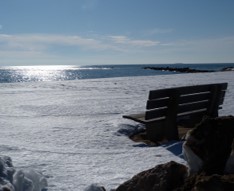}
        \end{minipage}
    \begin{minipage}{0.3\linewidth}
        \footnotesize{
        \textbf{GT:} A empty bench on a snow covered beach.\\
        \textbf{Transformer:} Two benches sitting on a beach near the water.\\
        \textbf{\ours:} A bench sitting on the beach in the snow.
        }
    \end{minipage}
    \hspace{0.2cm}
    \begin{minipage}{0.16\linewidth}
        \includegraphics[width=0.95\linewidth]{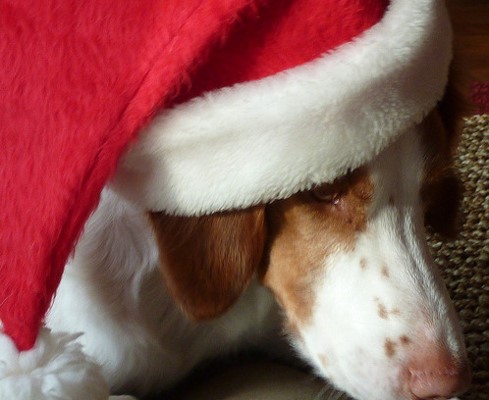}
        \end{minipage}
    \begin{minipage}{0.3\linewidth}
        \footnotesize{
        \textbf{GT:} A brown and white dog wearing a red and white Santa hat.\\
        \textbf{Transformer:} A white dog wearing a red hat.\\
        \textbf{\ours:} A dog wearing a red and white Santa hat.
        }
    \end{minipage}
                        
    \vspace{0.15cm}
    
    \begin{minipage}{0.16\linewidth}
        \includegraphics[width=0.95\linewidth]{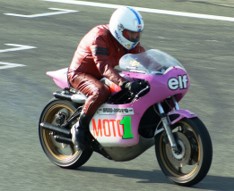}
        \end{minipage}
    \begin{minipage}{0.3\linewidth}
        \footnotesize{
        \textbf{GT:} A man riding a small pink motorcycle on a track.\\
        \textbf{Transformer:} A man is riding a red motorcycle.\\
        \textbf{\ours:} A man riding a pink motorcycle on a track.
        }
    \end{minipage}
    \hspace{0.2cm}
    \begin{minipage}{0.16\linewidth}
        \includegraphics[width=0.95\linewidth]{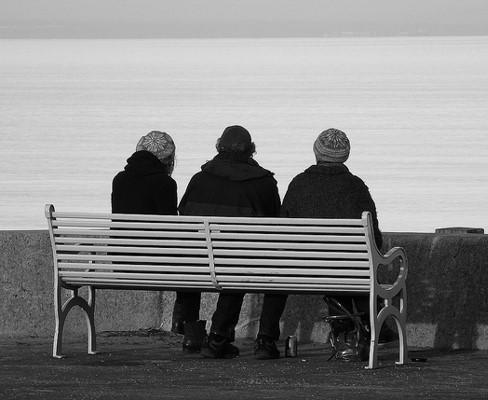}
        \end{minipage}
    \begin{minipage}{0.3\linewidth}
        \footnotesize{
        \textbf{GT:} Three people sit on a bench looking out over the water.\\
        \textbf{Transformer:} Two people sitting on a bench in the water.\\
        \textbf{\ours:} Three people sitting on a bench looking at the water.
        }
    \end{minipage}

\caption{Additional sample results generated by our approach and the original Transformer, as well as the corresponding ground-truths.}
\label{fig:results_supp1}
\vspace{-.3cm}
\end{figure*}

Figure~\ref{fig:results_supp1} shows additional qualitative results obtained from our model in comparison to the original Transformer and corresponding ground-truth captions. On average, the proposed model shows an improvement in terms of caption correctness and provides more detailed and exhaustive descriptions.

\begin{figure*}
    \centering
    \setlength{\tabcolsep}{.1em}
    \renewcommand*{\arraystretch}{0.5}
    \begin{tabular}{cccccc}

    \includegraphics[width=0.16\linewidth]{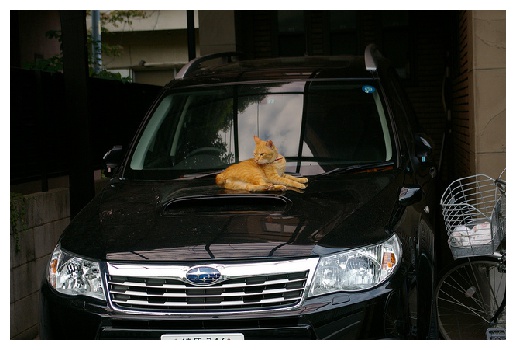} & 
    \includegraphics[width=0.16\linewidth]{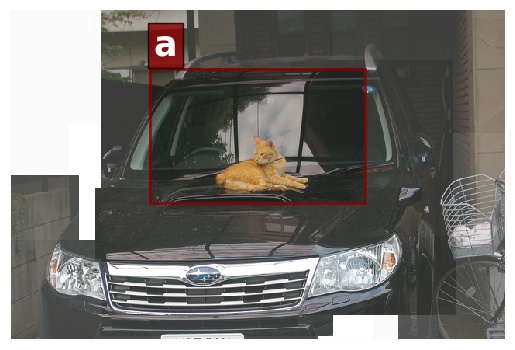} & 
    \includegraphics[width=0.16\linewidth]{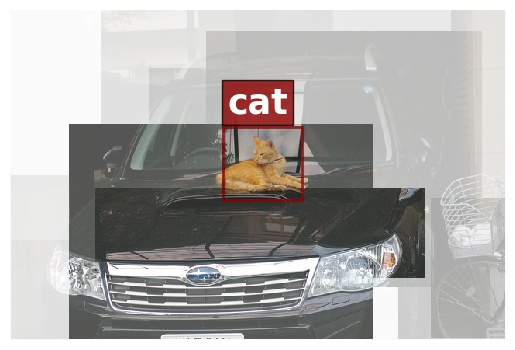} & 
    \includegraphics[width=0.16\linewidth]{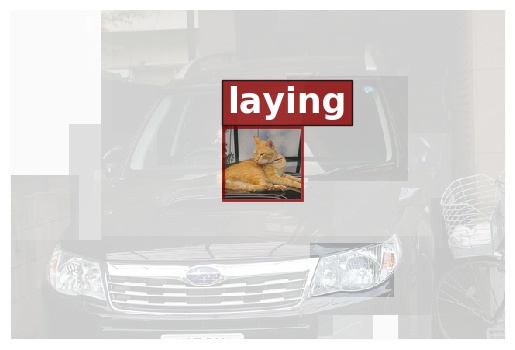} & 
    \includegraphics[width=0.16\linewidth]{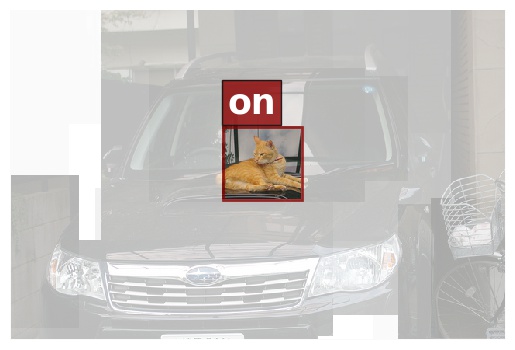} & 
    \includegraphics[width=0.16\linewidth]{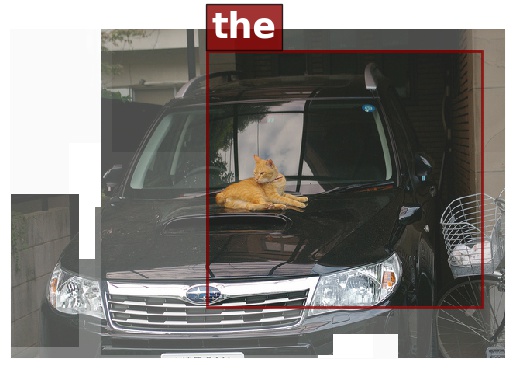} \\
    \includegraphics[width=0.16\linewidth]{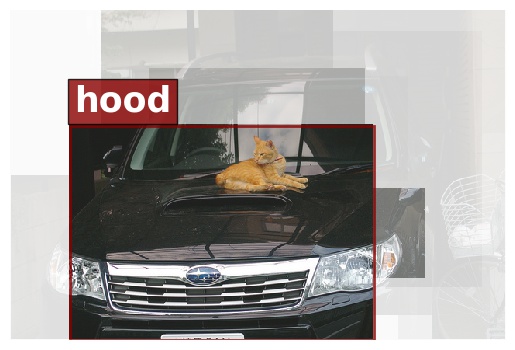} & 
    \includegraphics[width=0.16\linewidth]{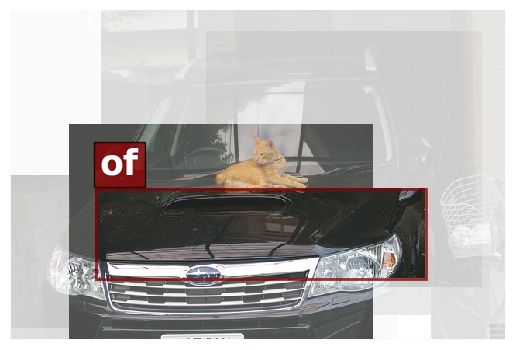} & 
    \includegraphics[width=0.16\linewidth]{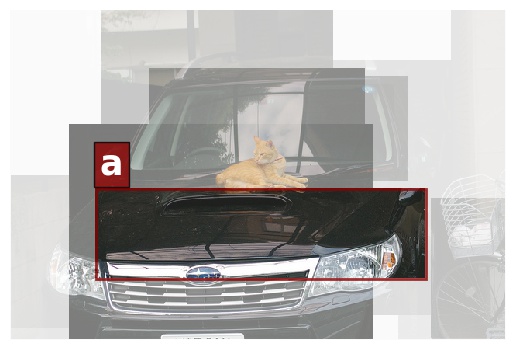} & 
    \includegraphics[width=0.16\linewidth]{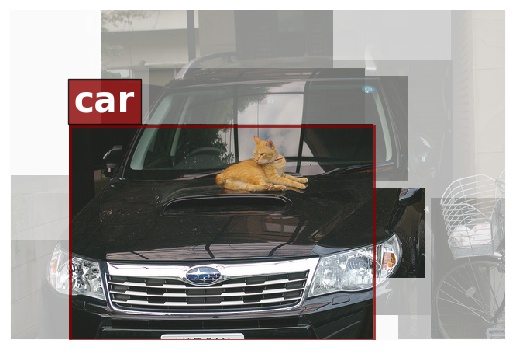} & 
     &
     \\
     \\
    \includegraphics[width=0.16\linewidth]{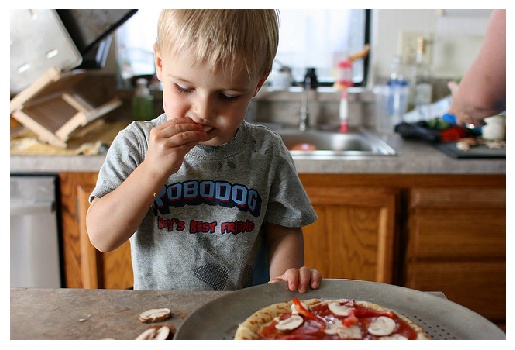} & 
    \includegraphics[width=0.16\linewidth]{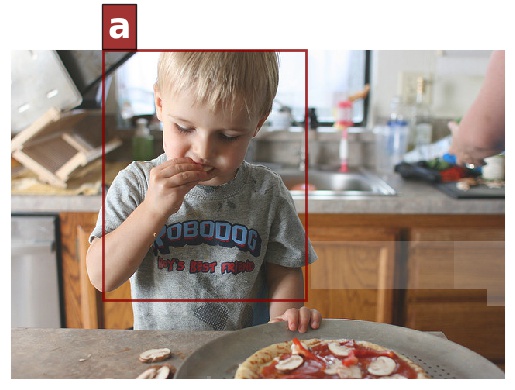} & 
    \includegraphics[width=0.16\linewidth]{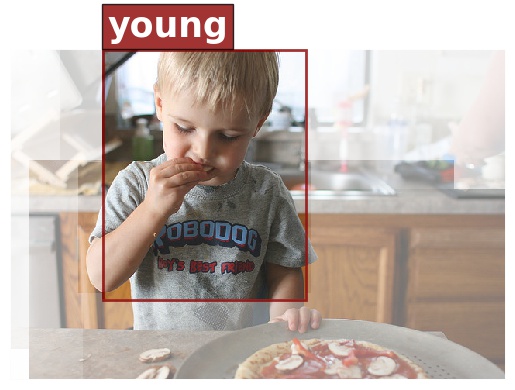} & 
    \includegraphics[width=0.16\linewidth]{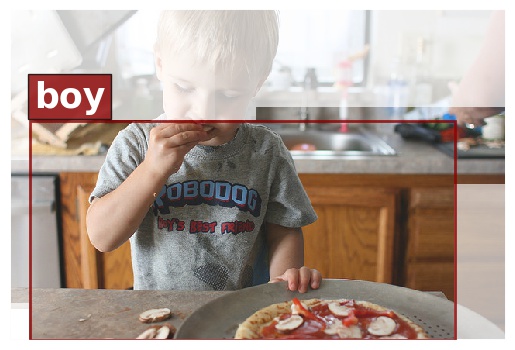} & 
    \includegraphics[width=0.16\linewidth]{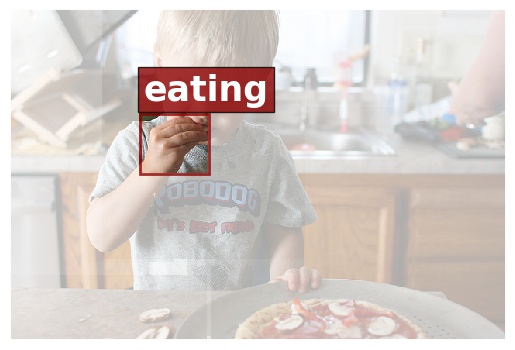} & 
    \includegraphics[width=0.16\linewidth]{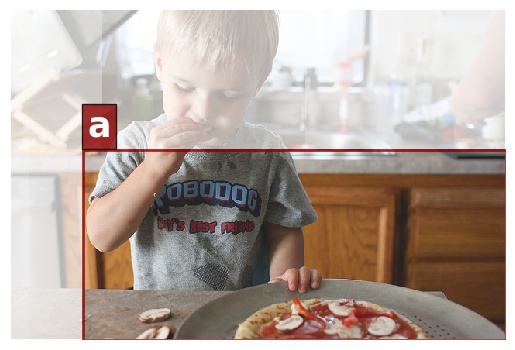} \\
    \includegraphics[width=0.16\linewidth]{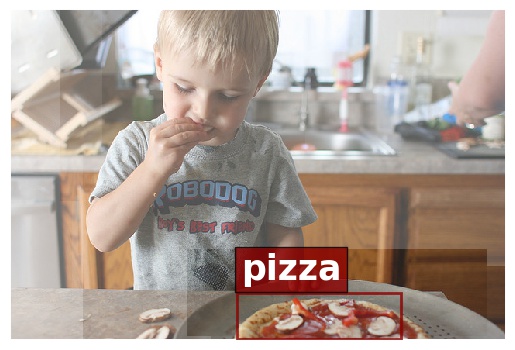} & 
    \includegraphics[width=0.16\linewidth]{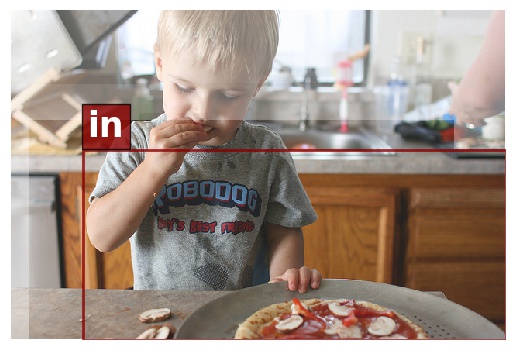} & 
    \includegraphics[width=0.16\linewidth]{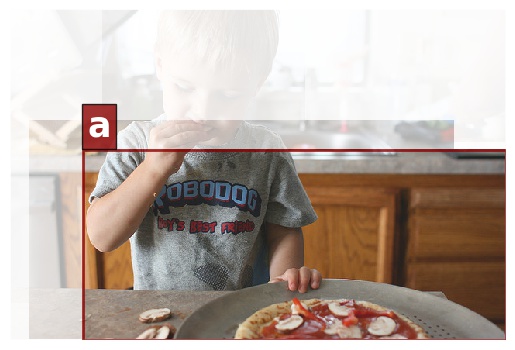} & 
    \includegraphics[width=0.16\linewidth]{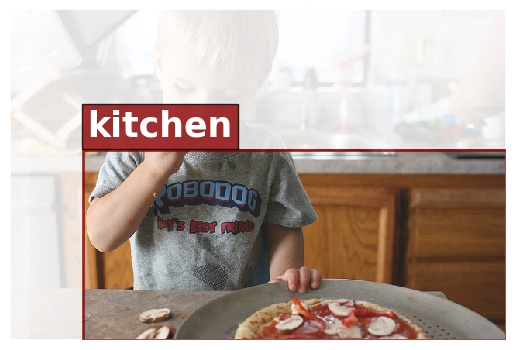} & 
     &
     \\
     \\
    \includegraphics[width=0.16\linewidth]{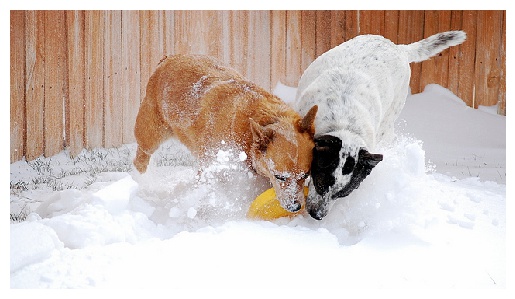} & 
    \includegraphics[width=0.16\linewidth]{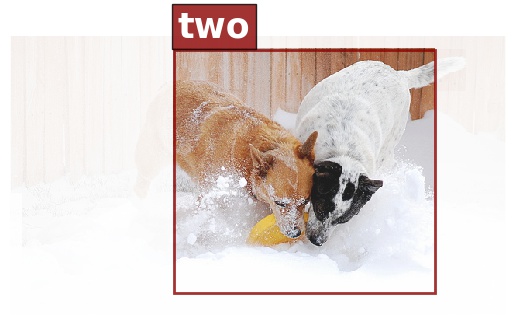} & 
    \includegraphics[width=0.16\linewidth]{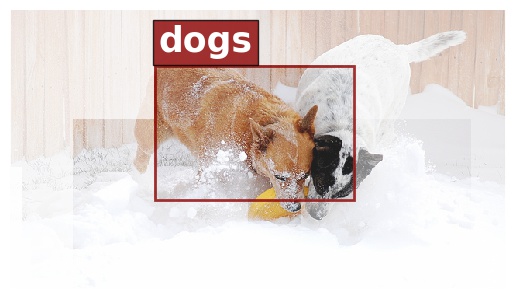} & 
    \includegraphics[width=0.16\linewidth]{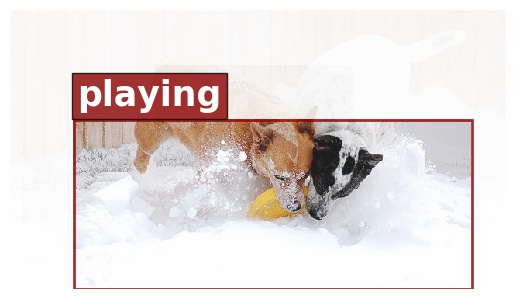} & 
    \includegraphics[width=0.16\linewidth]{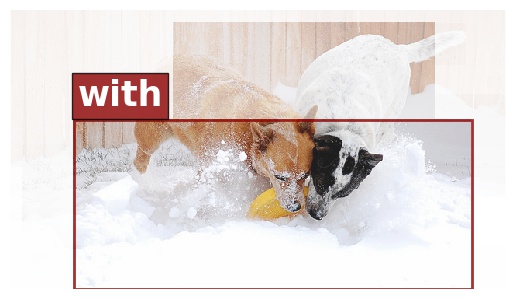} & 
    \includegraphics[width=0.16\linewidth]{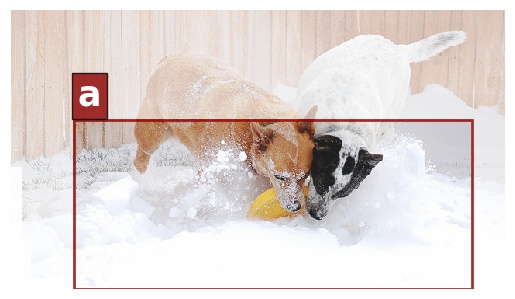} \\
    \includegraphics[width=0.16\linewidth]{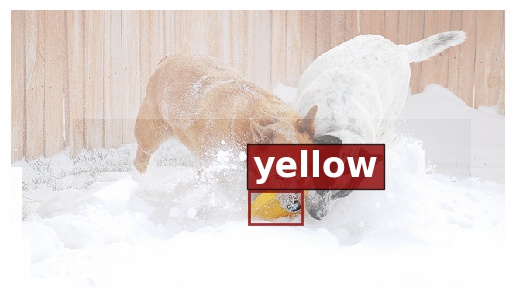} & 
    \includegraphics[width=0.16\linewidth]{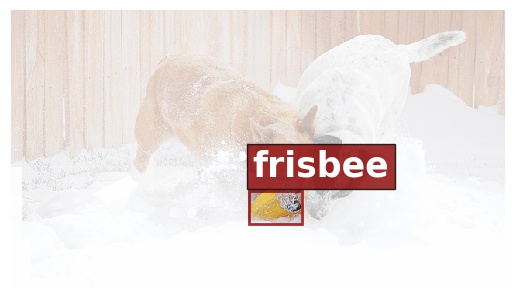} & 
    \includegraphics[width=0.16\linewidth]{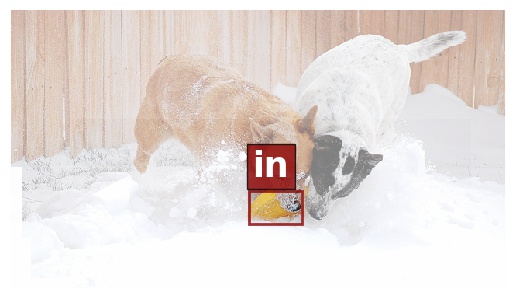} & 
    \includegraphics[width=0.16\linewidth]{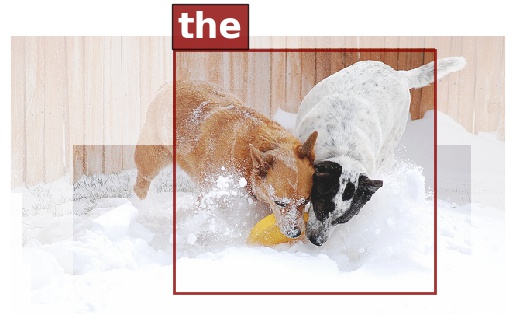} & 
    \includegraphics[width=0.16\linewidth]{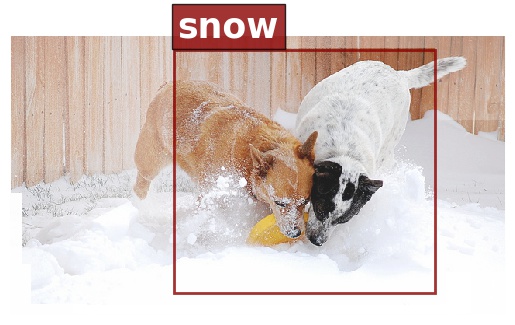} & 
     \\
     \\
    \includegraphics[width=0.125\linewidth]{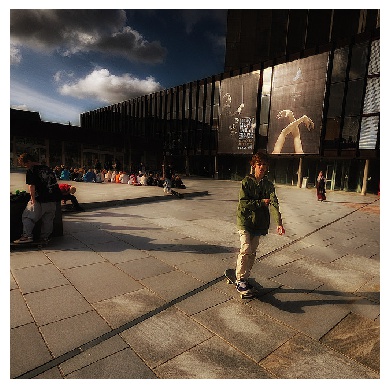} & 
    \includegraphics[width=0.125\linewidth]{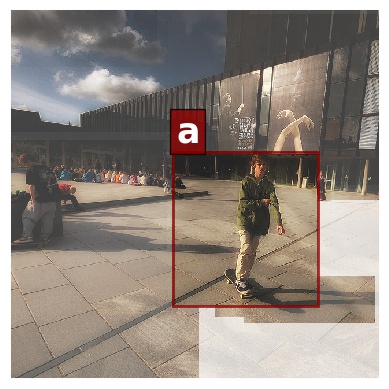} & 
    \includegraphics[width=0.125\linewidth]{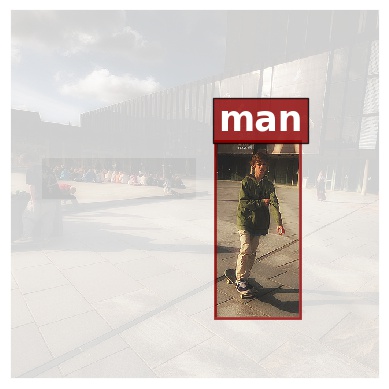} & 
    \includegraphics[width=0.125\linewidth]{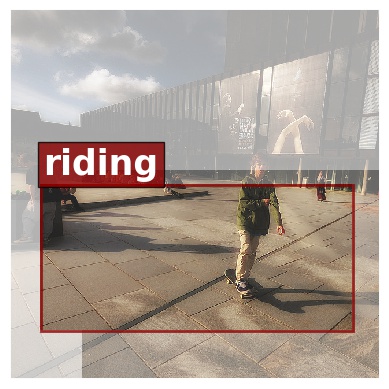} & 
    \includegraphics[width=0.125\linewidth]{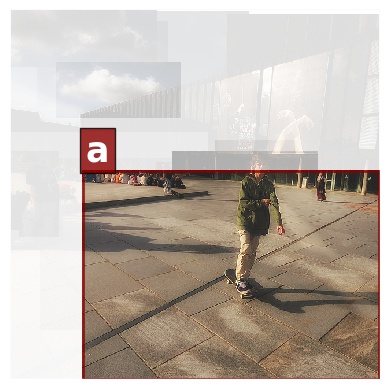} & 
    \includegraphics[width=0.125\linewidth]{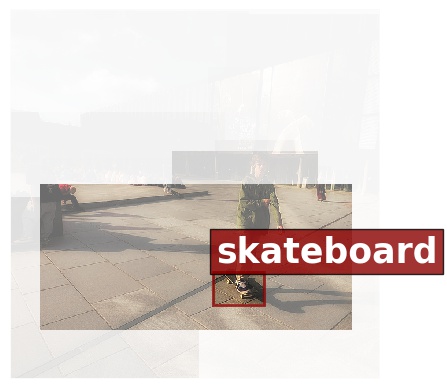} \\
    \includegraphics[width=0.125\linewidth]{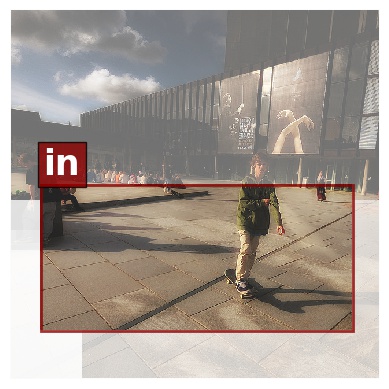} & 
    \includegraphics[width=0.125\linewidth]{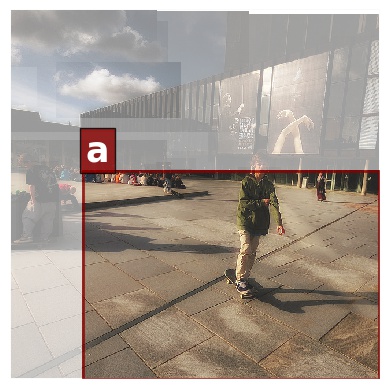} & 
    \includegraphics[width=0.125\linewidth]{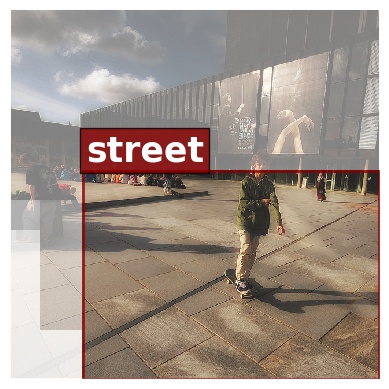} & 
     & 
     &
     \\
     \\
    \includegraphics[width=0.16\linewidth]{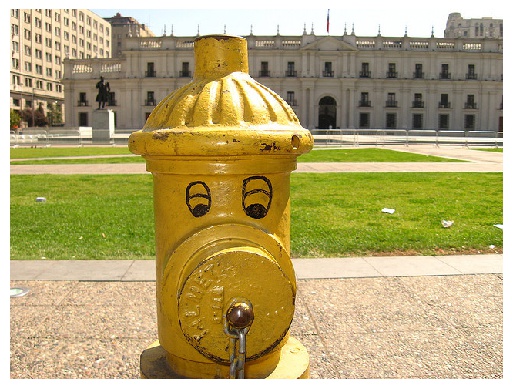} & 
    \includegraphics[width=0.16\linewidth]{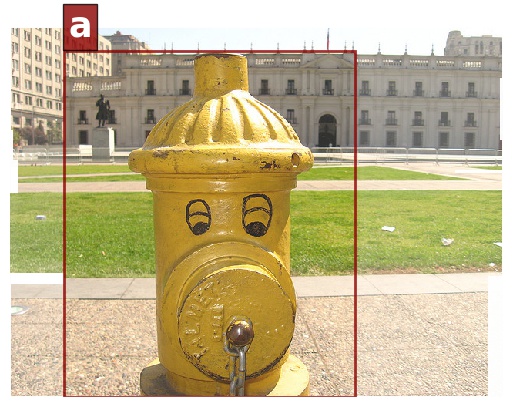} & 
    \includegraphics[width=0.16\linewidth]{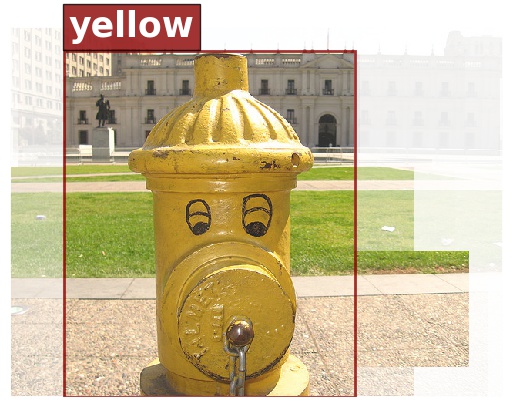} & 
    \includegraphics[width=0.16\linewidth]{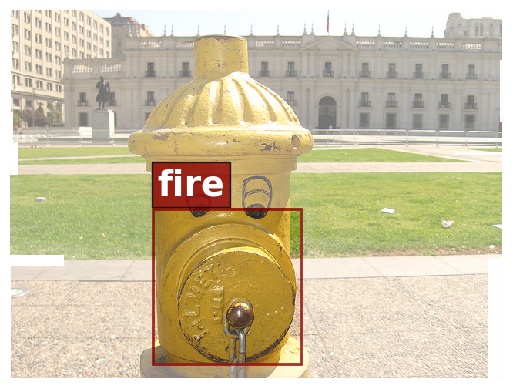} & 
    \includegraphics[width=0.16\linewidth]{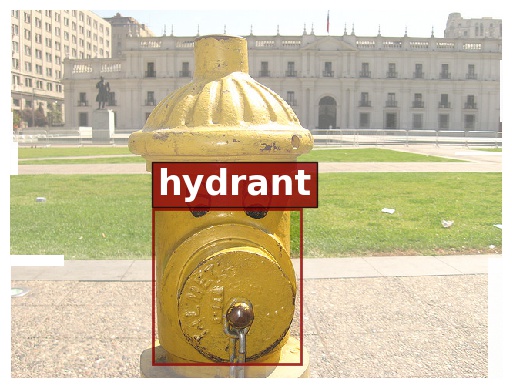} & 
    \includegraphics[width=0.16\linewidth]{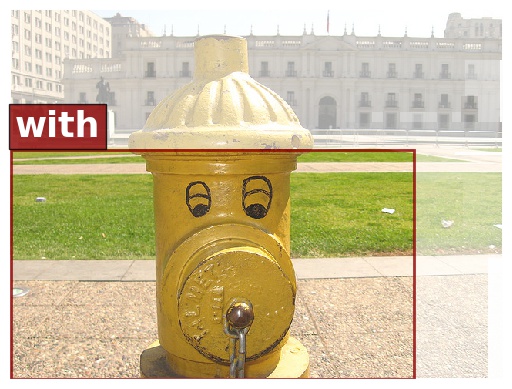} \\
    \includegraphics[width=0.16\linewidth]{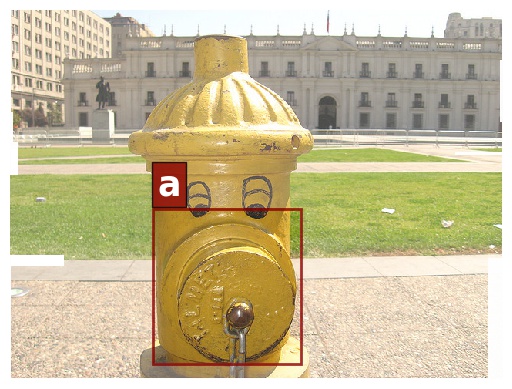} & 
    \includegraphics[width=0.16\linewidth]{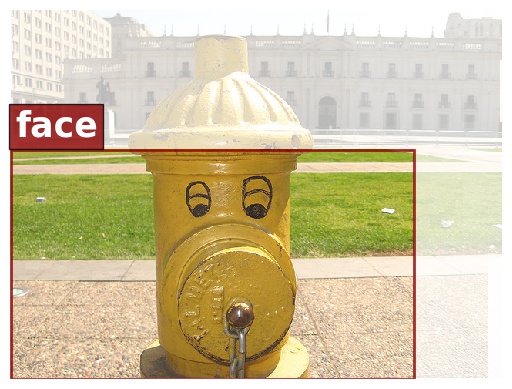} & 
    \includegraphics[width=0.16\linewidth]{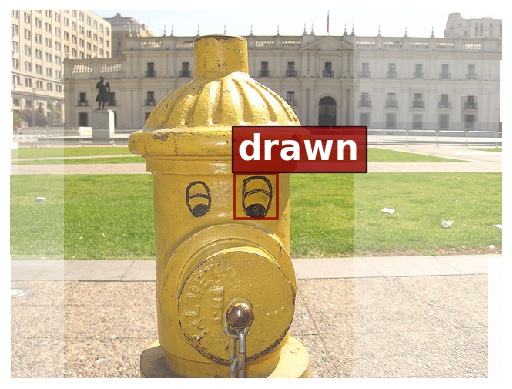} & 
    \includegraphics[width=0.16\linewidth]{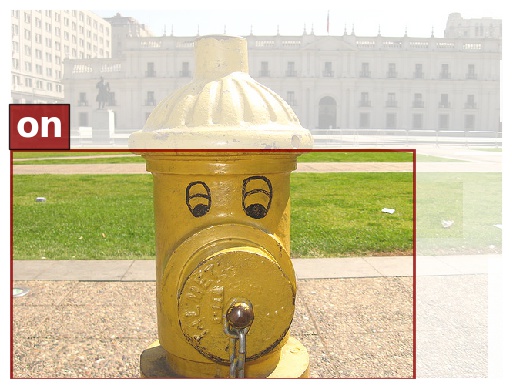} & 
    \includegraphics[width=0.16\linewidth]{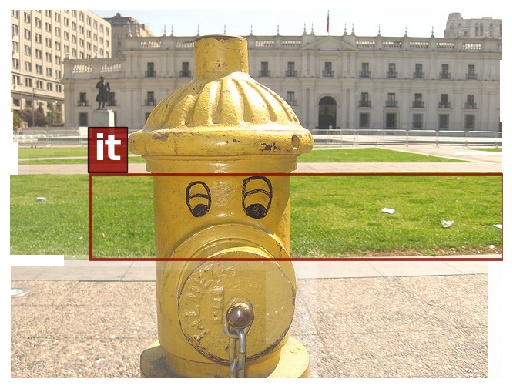} &
     \\
    \end{tabular}
    \caption{Visualization of attention states for sample captions generated by our~\ours. For each generated word, we show the attended image regions, outlining the region with the maximum output attribution in red.}
    \label{fig:attention1}
\vspace{-.3cm}
\end{figure*}

\begin{figure*}
    \centering
    \setlength{\tabcolsep}{.1em}
    \renewcommand*{\arraystretch}{0.5}
    \begin{tabular}{cccccc}

    \includegraphics[width=0.16\linewidth]{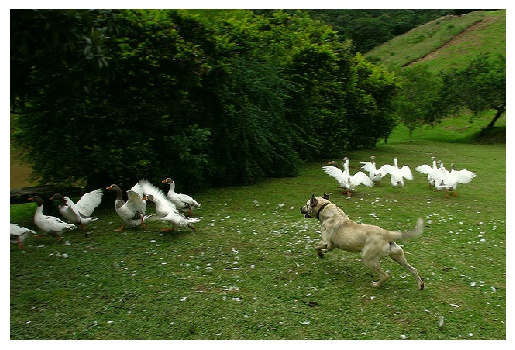} & 
    \includegraphics[width=0.16\linewidth]{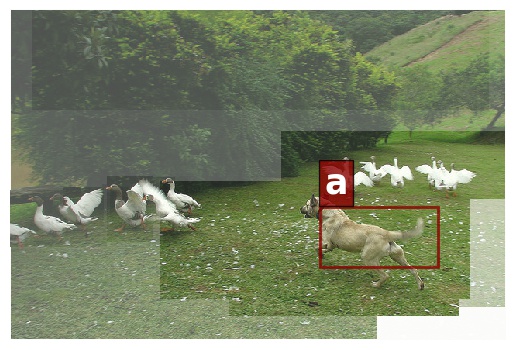} & 
    \includegraphics[width=0.16\linewidth]{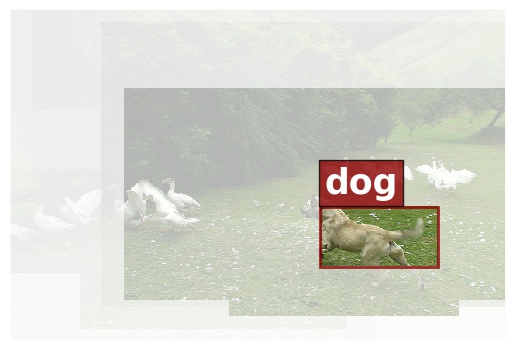} & 
    \includegraphics[width=0.16\linewidth]{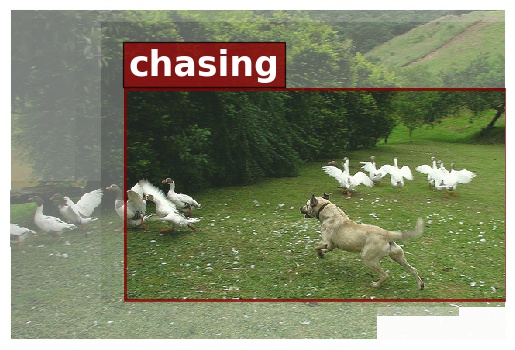} & 
    \includegraphics[width=0.16\linewidth]{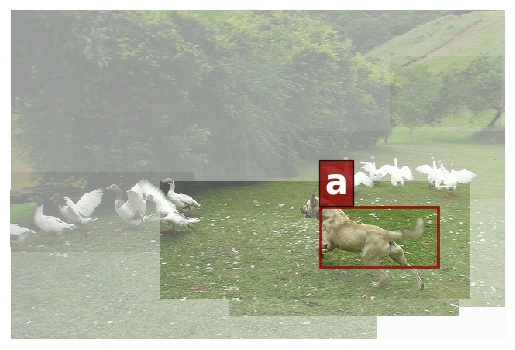} & 
    \includegraphics[width=0.16\linewidth]{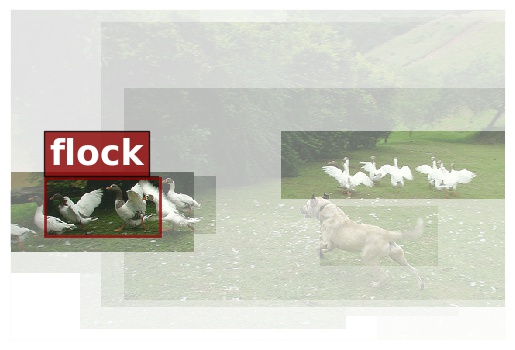} \\
    \includegraphics[width=0.16\linewidth]{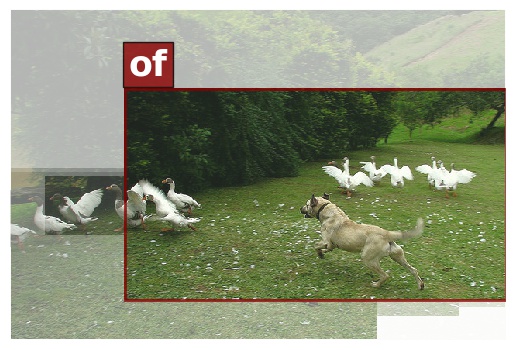} & 
    \includegraphics[width=0.16\linewidth]{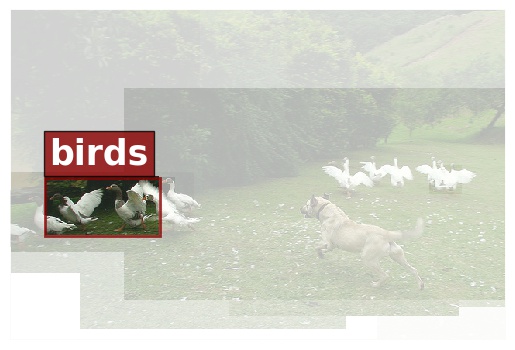} & 
    \includegraphics[width=0.16\linewidth]{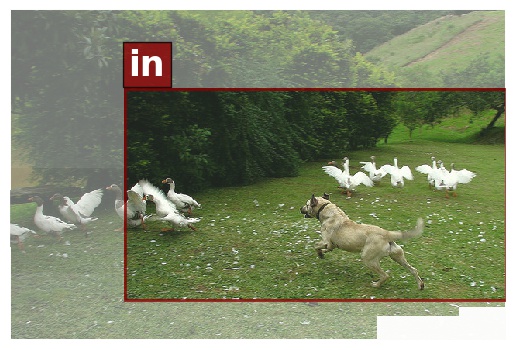} & 
    \includegraphics[width=0.16\linewidth]{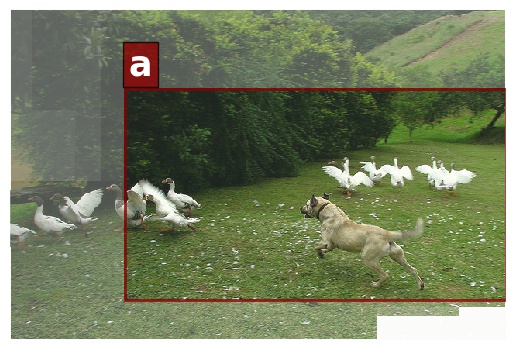} & 
    \includegraphics[width=0.16\linewidth]{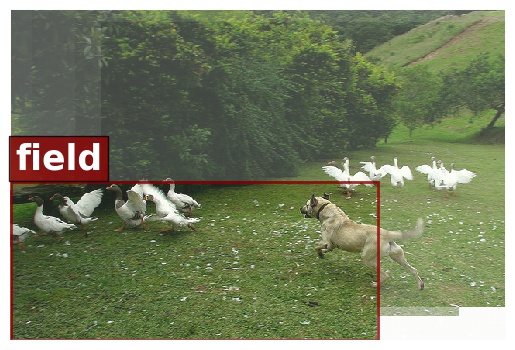} & 
     \\
     \\
    \includegraphics[width=0.16\linewidth]{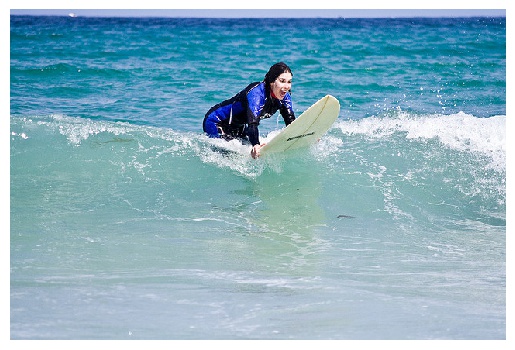} & 
    \includegraphics[width=0.16\linewidth]{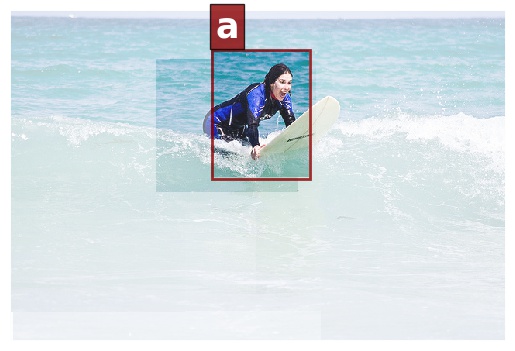} & 
    \includegraphics[width=0.16\linewidth]{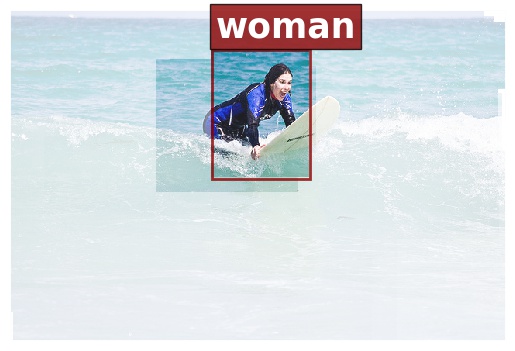} & 
    \includegraphics[width=0.16\linewidth]{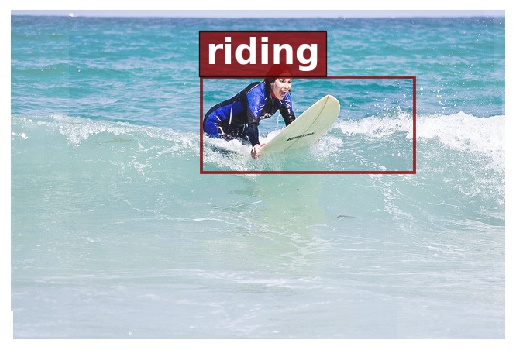} & 
    \includegraphics[width=0.16\linewidth]{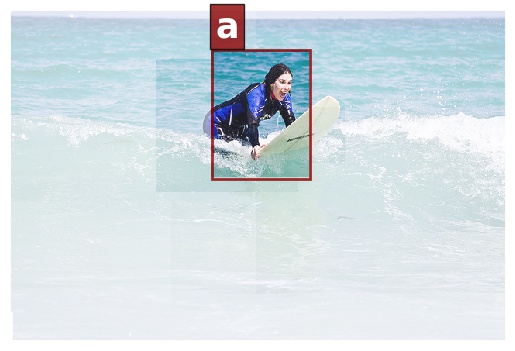} & 
    \includegraphics[width=0.16\linewidth]{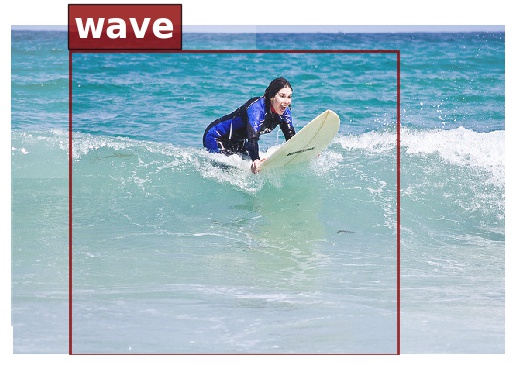} \\
    \includegraphics[width=0.16\linewidth]{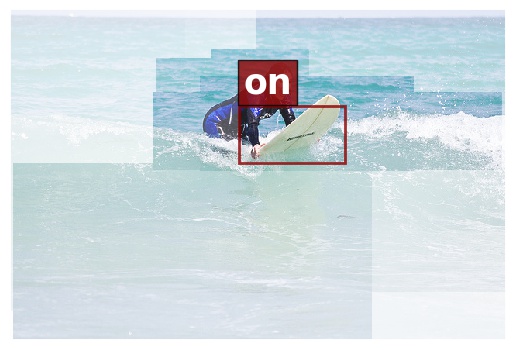} & 
    \includegraphics[width=0.16\linewidth]{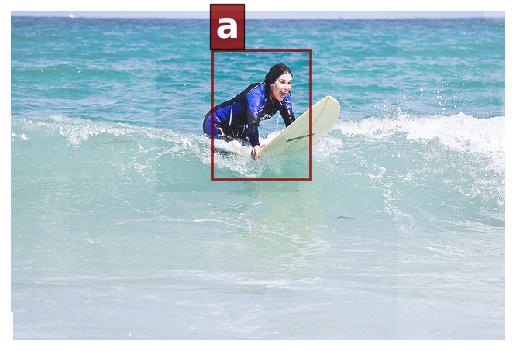} & 
    \includegraphics[width=0.16\linewidth]{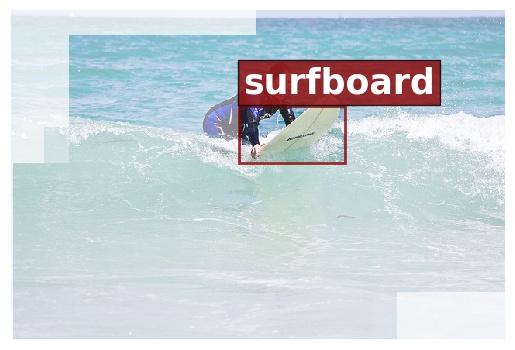} & 
    \includegraphics[width=0.16\linewidth]{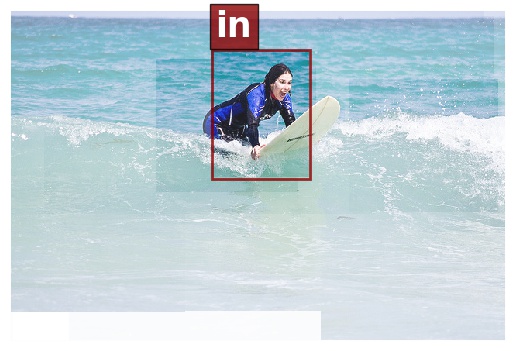} & 
    \includegraphics[width=0.16\linewidth]{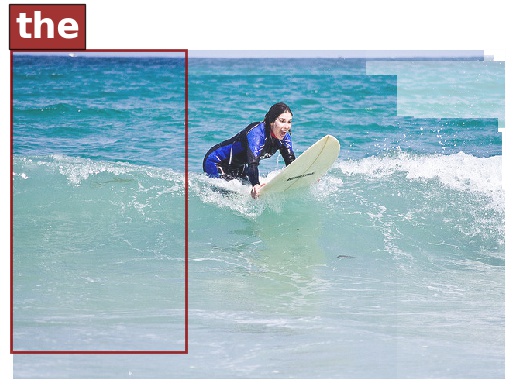} &
    \includegraphics[width=0.16\linewidth]{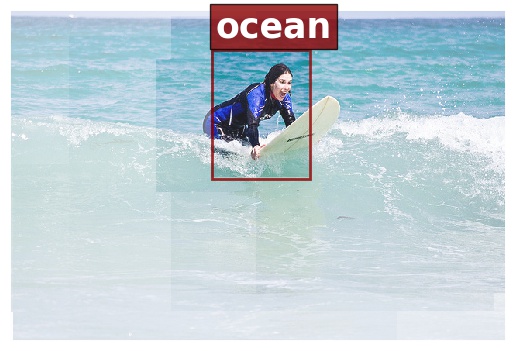} \\
     \\
    \includegraphics[width=0.16\linewidth]{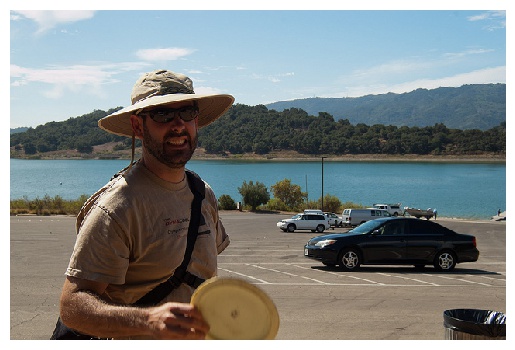} & 
    \includegraphics[width=0.16\linewidth]{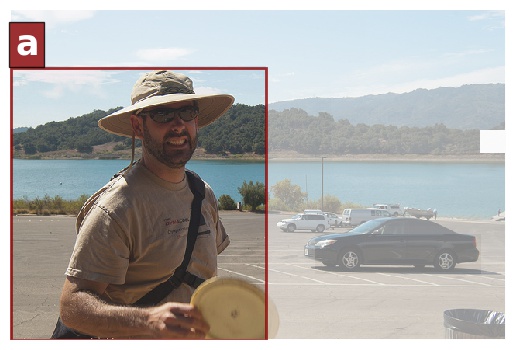} & 
    \includegraphics[width=0.16\linewidth]{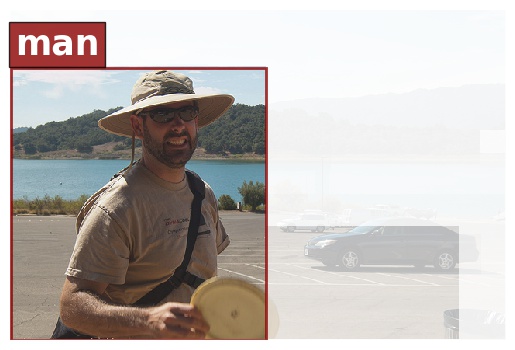} & 
    \includegraphics[width=0.16\linewidth]{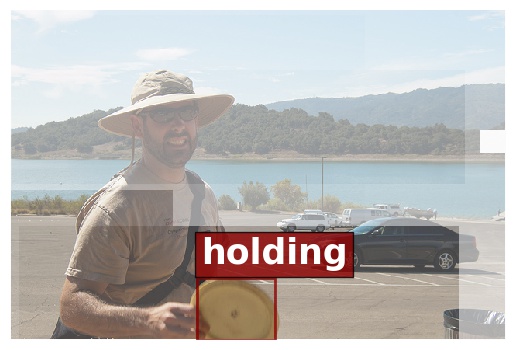} & 
    \includegraphics[width=0.16\linewidth]{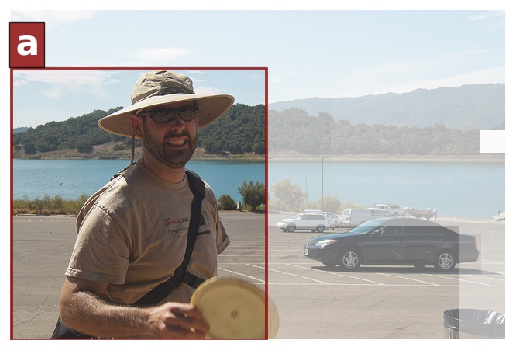} & 
    \includegraphics[width=0.16\linewidth]{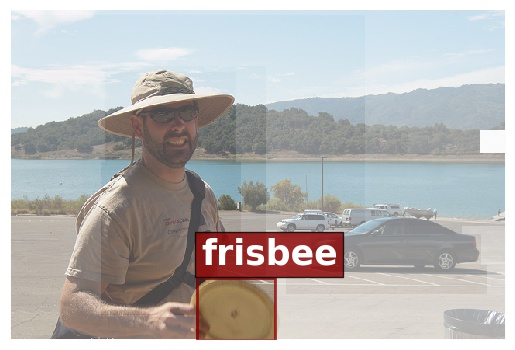} \\
    \includegraphics[width=0.16\linewidth]{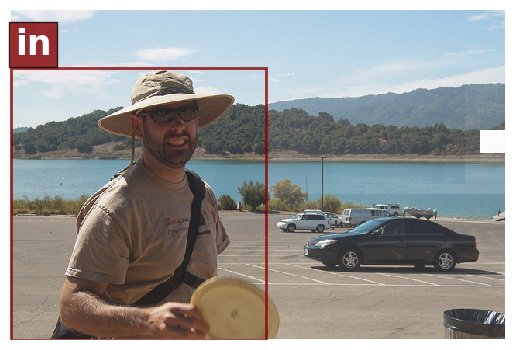} & 
    \includegraphics[width=0.16\linewidth]{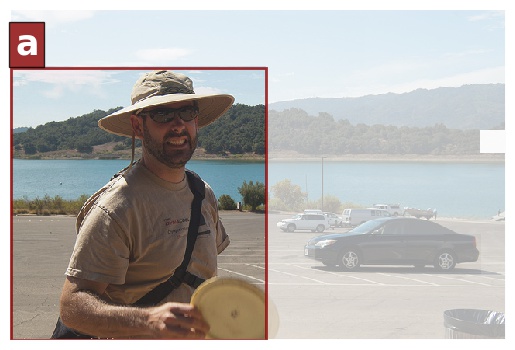} & 
    \includegraphics[width=0.16\linewidth]{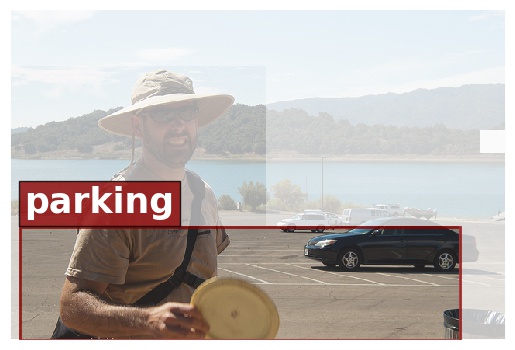} & 
    \includegraphics[width=0.16\linewidth]{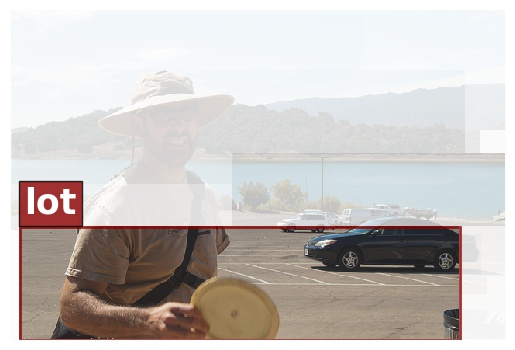} & 
    & 
     \\
     \\
    \includegraphics[width=0.16\linewidth]{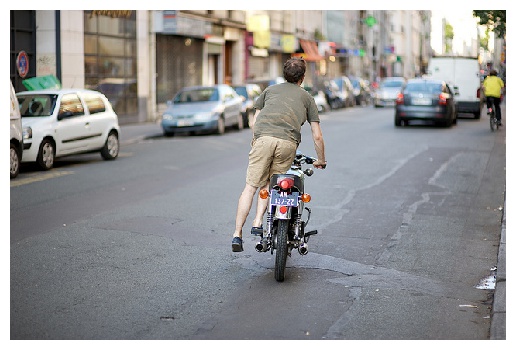} & 
    \includegraphics[width=0.16\linewidth]{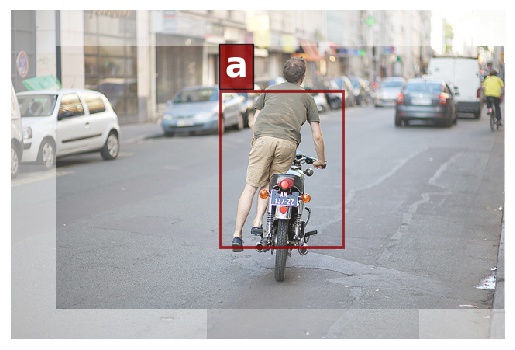} & 
    \includegraphics[width=0.16\linewidth]{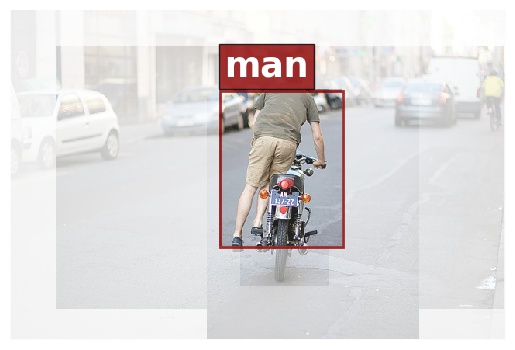} & 
    \includegraphics[width=0.16\linewidth]{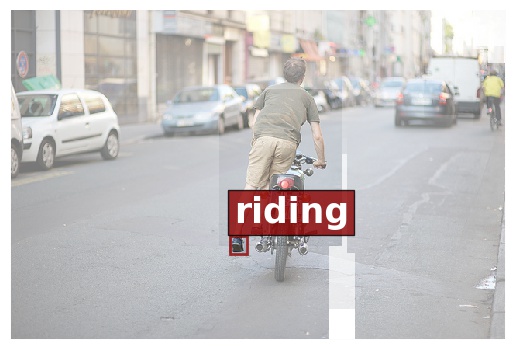} & 
    \includegraphics[width=0.16\linewidth]{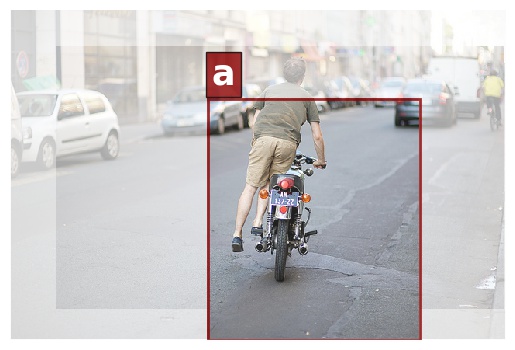} & 
    \includegraphics[width=0.16\linewidth]{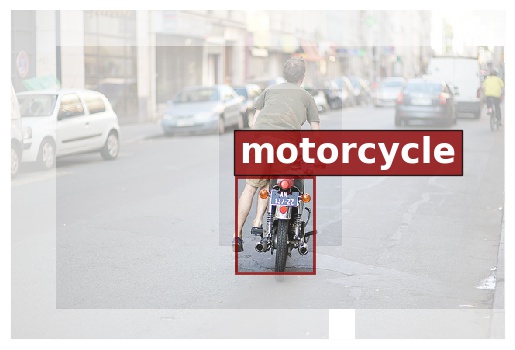} \\
    \includegraphics[width=0.16\linewidth]{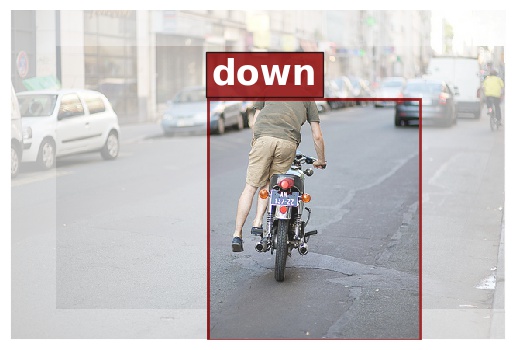} & 
    \includegraphics[width=0.16\linewidth]{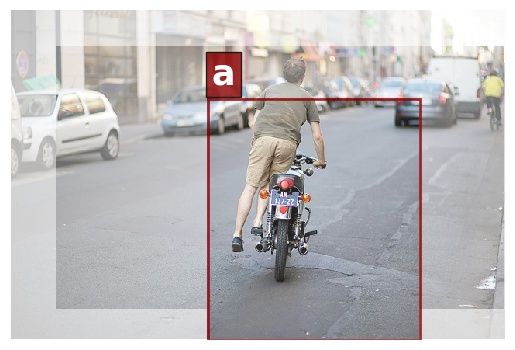} & 
    \includegraphics[width=0.16\linewidth]{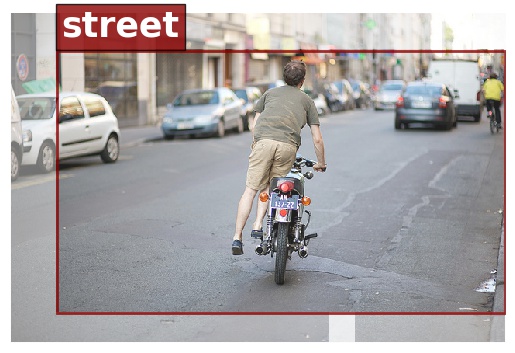} & 
     & 
     &
     \\
     \\
    \includegraphics[width=0.16\linewidth]{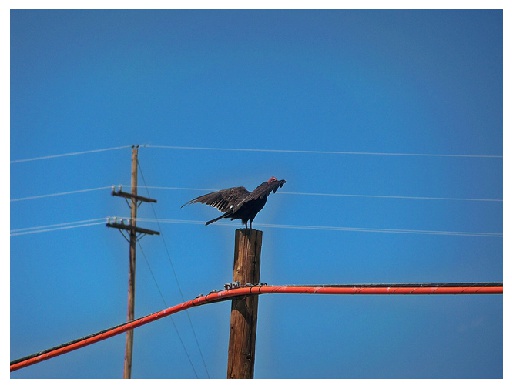} & 
    \includegraphics[width=0.16\linewidth]{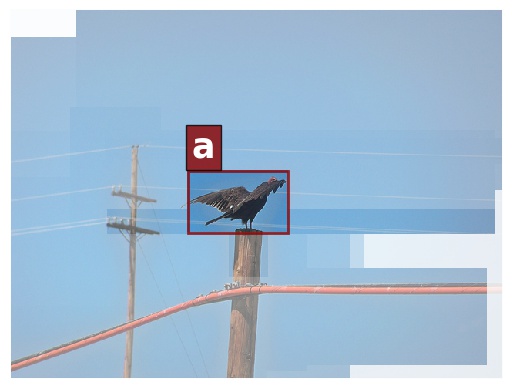} & 
    \includegraphics[width=0.16\linewidth]{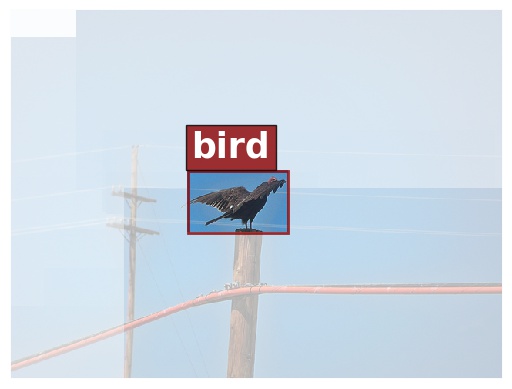} & 
    \includegraphics[width=0.16\linewidth]{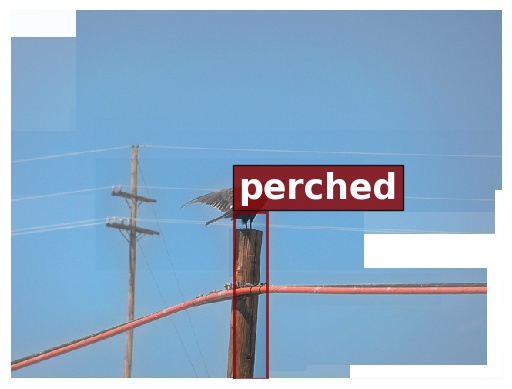} & 
    \includegraphics[width=0.16\linewidth]{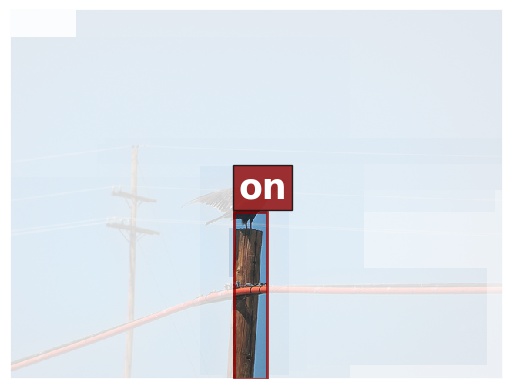} & 
    \includegraphics[width=0.16\linewidth]{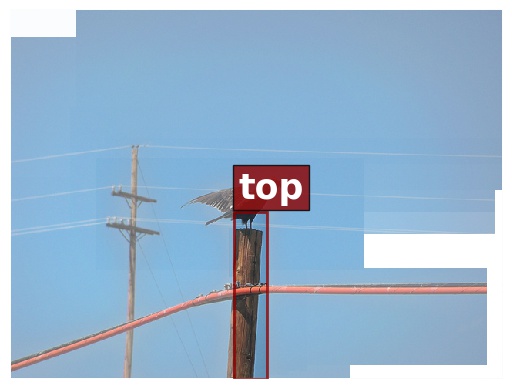} \\
    \includegraphics[width=0.16\linewidth]{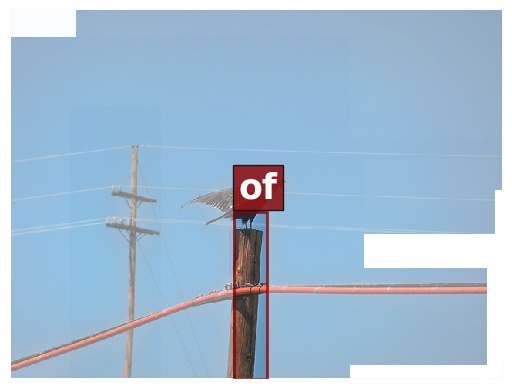} & 
    \includegraphics[width=0.16\linewidth]{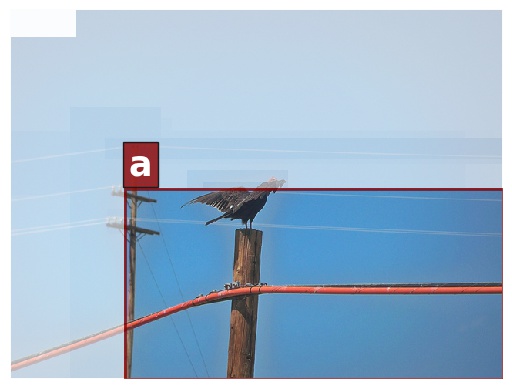} & 
    \includegraphics[width=0.16\linewidth]{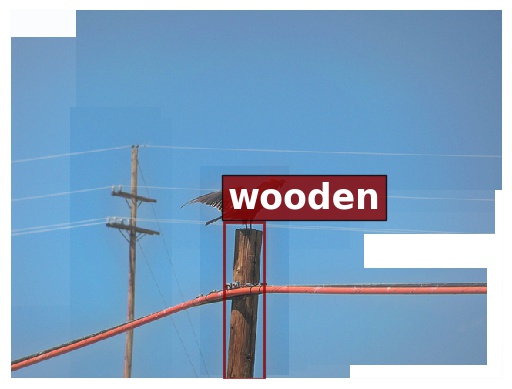} & 
    \includegraphics[width=0.16\linewidth]{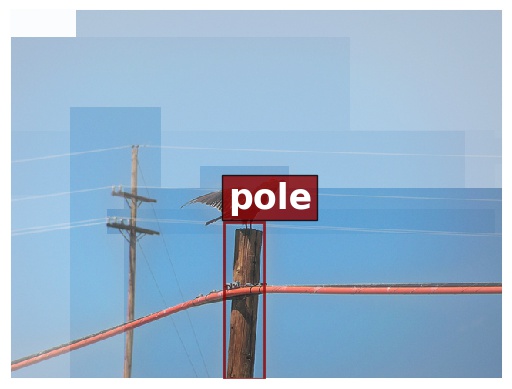} & 
     &
     \\
    \end{tabular}
    \caption{Visualization of attention states for sample captions generated by our~\ours. For each generated word, we show the attended image regions, outlining the region with the maximum output attribution in red.}
    \label{fig:attention2}
\vspace{-.3cm}
\end{figure*}

Figures~\ref{fig:attention1} and~\ref{fig:attention2}, instead, report the visualization of attentive states on a variety of sample images, following the approach outlined in Sec.~4.6 of the main paper. Specifically, the Integrated Gradients approach~\cite{sundararajan2017axiomatic} produces an attribution score for each feature channel of each input region. To obtain the attribution of each region, we average over the feature channels, and re-normalize the obtained scores by their sum. For visualization purposes, we apply a contrast stretching function to project scores in the 0-1 interval.

\subsection{Novel object captioning} 
Figure~\ref{fig:nocaps} reports sample captions produced by our approach on images from the nocaps dataset. On each image, we compare to the baseline Transformer and show the constraints provided by the object detector. Overall, the \ours is able to better incorporate the constraints while maintaining the fluency and properness of the generated sentences.

Following~\cite{agrawal2019nocaps}, we use an object detector trained on Open Images~\footnote{Specifically, the \scriptsize{\texttt{tf\_faster\_rcnn\_inception\_resnet\_v2\_atrous\_oidv2}} \footnotesize{model from the Tensorflow model zoo.}} and filter detections by removing 39 Open Images classes that contain parts of objects or which are seldom mentioned. We also discard overlapping detections by removing the higher-order of two objects based on the class hierarchy, and we use the top-3 detected objects as constraints based on the detection confidence score. As mentioned in Sec.~4.5 and differently from~\cite{agrawal2019nocaps}, we do not consider the plural forms or other word phrases of object classes, thus taking into account only the original class names. After decoding, we select the predicted caption with highest probability that satisfies the given constraints. 

\end{document}